\documentclass[10pt,twocolumn,letterpaper]{article}

\usepackage[pagenumbers]{cvpr} % To force page numbers, e.g. for an arXiv version

\definecolor{cvprblue}{rgb}{0.21,0.49,0.74}
\usepackage[pagebackref,breaklinks,colorlinks,allcolors=cvprblue]{hyperref}

\usepackage{multirow}

\def\paperID{5301} % *** Enter the Paper ID here
\def\confName{CVPR}
\def\confYear{2025}

\title{BiDense: Binarization for Dense Prediction}

\author{
Rui Yin$^1$
\qquad
Haotong Qin$^2$
\qquad
Yulun Zhang$^3$
\qquad
Wenbo Li$^4$
\\
Yong Guo$^5$
\qquad
Jianjun Zhu$^1$
\qquad
Cheng Wang$^1$
\qquad
Biao Jia$^1$\thanks{corresponding author}
\\
$^1$Hanglok-Tech
\qquad
$^2$ETH Zürich
\qquad
$^3$Shanghai Jiao Tong University
\\
$^4$The Chinese University of Hong Kong
\qquad
$^5$South China University of Technology
}

\begin{document}
\maketitle
\begin{abstract}
Dense prediction is a critical task in computer vision. 
However, previous methods often require extensive computational resources, which hinders their real-world application. 
In this paper, we propose BiDense, a generalized binary neural network (BNN) designed for efficient and accurate dense prediction tasks. 
BiDense incorporates two key techniques: the Distribution-adaptive Binarizer (DAB) and the Channel-adaptive Full-precision Bypass (CFB).
The DAB adaptively calculates thresholds and scaling factors for binarization, effectively retaining more information within BNNs. 
Meanwhile, the CFB facilitates full-precision bypassing for binary convolutional layers undergoing various channel size transformations, which enhances the propagation of real-valued signals and minimizes information loss.
By leveraging these techniques, BiDense preserves more real-valued information, enabling more accurate and detailed dense predictions in BNNs. 
Extensive experiments demonstrate that our framework achieves performance levels comparable to full-precision models while significantly reducing memory usage and computational costs.
Project code: \hyperlink{https://github.com/frickyinn/BiDense}{https://github.com/frickyinn/BiDense}
\end{abstract}    
\vspace{-1mm}
\section{Introduction}
\vspace{-1mm}
\label{sec:intro}

Dense prediction involves estimating target results for every pixel in an image, including classification tasks (e.g., semantic segmentation), regression tasks (e.g., monocular depth estimation), etc.
Traditionally, neural networks designed for dense prediction follow a UNet-like architecture~\cite{ronneberger2015u}, which comprises an encoder and a decoder.
% The encoder enlarges the receptive field and increases the number of feature channels, extracting deeper image information. 
% In contrast, the decoder restores the image dimensions while reducing channel sizes, producing fine-grained predictions.
Existing methods utilize convolutional neural networks (CNNs) or transformers, employing more parameters and computational operations for more accurate and detailed estimations.
However, the increasing demand for computational resources creates a significant barrier to real-world applications, particularly on resource-constrained edge devices.

\begin{figure}[htbp]
    \centering
    \footnotesize
    \tabcolsep=1.5pt
    \begin{tabular}{cccc}
        & ConvNeXt~\cite{liu2022convnet} & BiDense (Ours) & Ground Truth \\
        \includegraphics[width=0.235\linewidth]{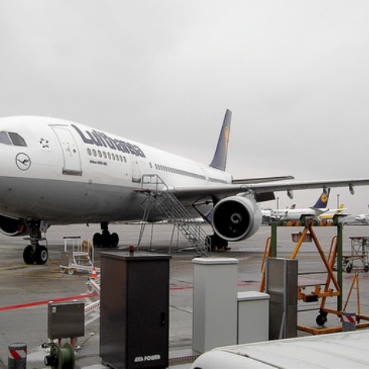} & 
        \includegraphics[width=0.235\linewidth]{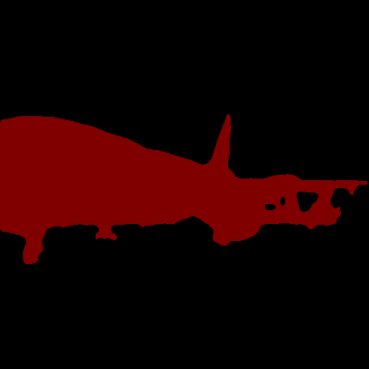} & 
        \includegraphics[width=0.235\linewidth]{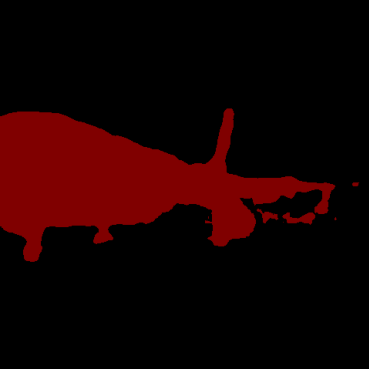} &
        \includegraphics[width=0.235\linewidth]{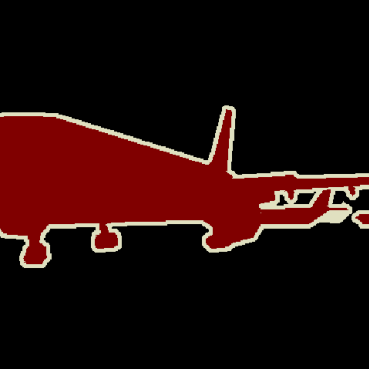} \\
        
        \includegraphics[width=0.235\linewidth]{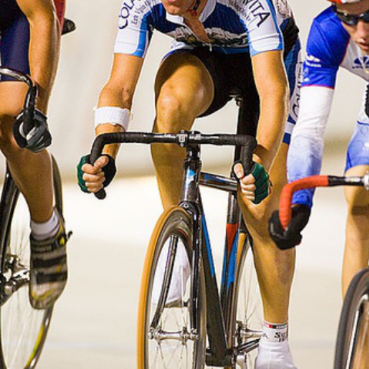} & 
        \includegraphics[width=0.235\linewidth]{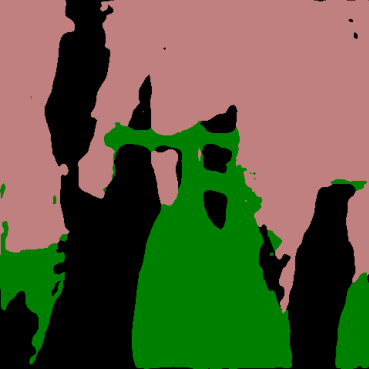} & 
        \includegraphics[width=0.235\linewidth]{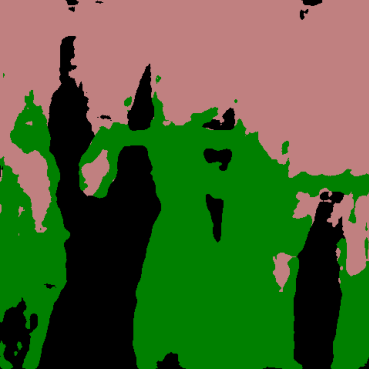} &
        \includegraphics[width=0.235\linewidth]{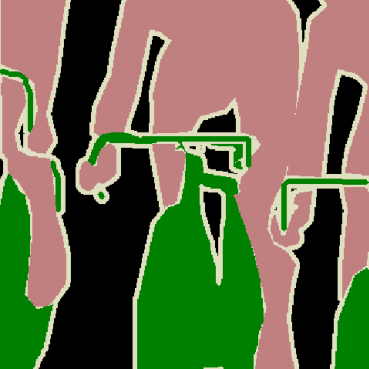} \\
        
        \includegraphics[width=0.235\linewidth]{fig/qualitative_results/nyu/img_400.png} & 
        \includegraphics[width=0.235\linewidth]{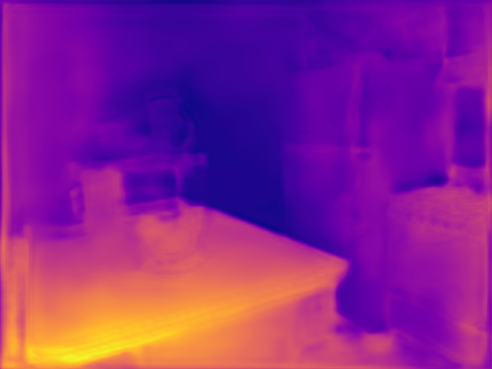} & 
        \includegraphics[width=0.235\linewidth]{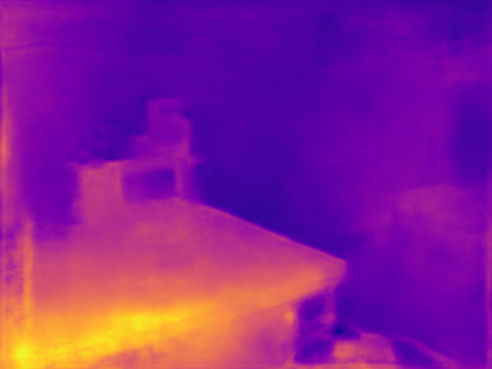} &
        \includegraphics[width=0.235\linewidth]{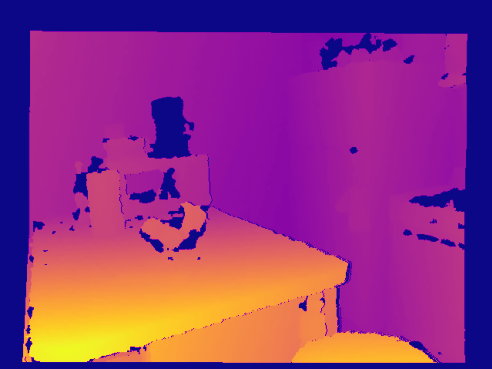} \\
        
        \includegraphics[width=0.235\linewidth]{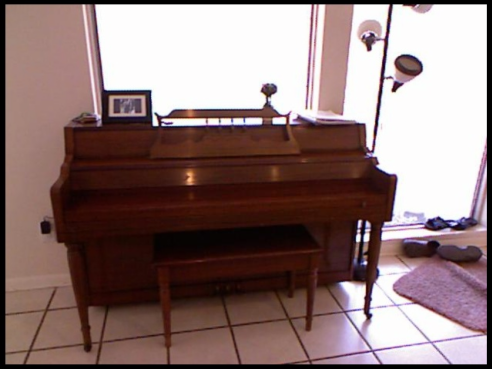} & 
        \includegraphics[width=0.235\linewidth]{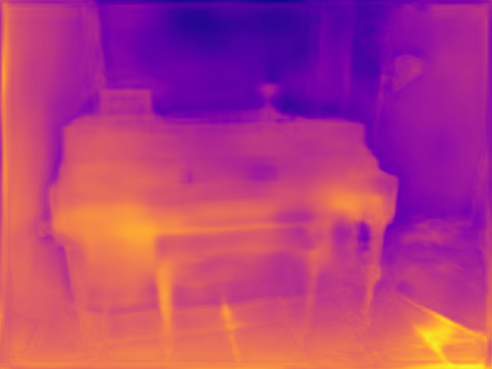} & 
        \includegraphics[width=0.235\linewidth]{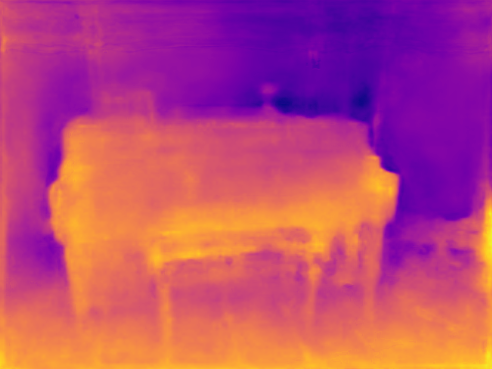} &
        \includegraphics[width=0.235\linewidth]{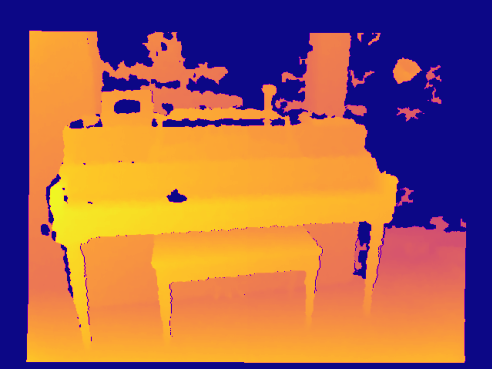} \\
    \end{tabular}
    \vspace{-2mm}
    \caption{\textbf{BiDense} approaches the performance of full-precision 32-bit networks in dense prediction tasks while significantly reducing memory and computational costs.}
    \vspace{-4mm}
    \label{fig:teaser}
\end{figure}

To compress and accelerate neural networks, many methods have been proposed~\cite{sze2017efficient, liu2018bi, wangtrainability, li2017pruning, hinton2015distilling, zhang2018shufflenet, ma2018shufflenet, howard2017mobilenets, sandler2018mobilenetv2}.
Among these, model quantization~\cite{hubara2016binarized, courbariaux2015binaryconnect} stands out for its ability to compress the weights and activations of networks into low-bit representations. 
As the most extreme form, model binarization restricts all variables to binary values, resulting in binary neural networks (BNNs).
By using only 1-bit representations, BNNs achieve a memory compression ratio of up to $32\times$ and a computational reduction of $58\times$ compared to full-precision 32-bit networks.
Although model binarization offers substantial resource savings, it often leads to severe performance degradation. 
As the real-valued variables are binarized, BNNs lose full-precision information, causing drops in accuracy. 
To mitigate this issue, various methods~\cite{liu2020reactnet, rastegari2016xnor, liu2022bit, tu2022adabin, he2023bivit, cai2024binarized} have been proposed to retain real-valued information during the binarization process. 
These methods introduce a limited number of full-precision learnable parameters to incorporate binarized variables with real-valued information.
However, since the parameters are fixed after the training stage, they only achieve optimal solutions on the training set. 
Consequently, they may exhibit inferior performance when faced with unseen cases or fluctuating outlier inputs.

Although some BNNs~\cite{zhuang2019structured, zhou2022cellular} have focused on resource-intensive dense prediction tasks, they primarily target semantic segmentation and enhancing representation capability. 
However, the critical challenge for generalized pixel-level estimation is minimizing information loss and effectively incorporating fine-grained features in BNNs to yield more detailed and accurate results.
Previous work~\cite{cai2024binarized} has proposed specialized binary convolutional layers that allow for full-precision bypasses when the input and output channels mismatch, propagating real-valued fine-grained signals. 
Yet, it is only applicable when the channel size is reduced by half or increased by double, ineffective in many other situations.
Therefore, there is an urgent need to develop generalized techniques that facilitate real-valued skip connections across all channel transformations, thereby minimizing information loss.

To address the aforementioned challenges, we propose a \textbf{Bi}narization framework for generalized \textbf{Dense} prediction tasks (\textbf{BiDense}). 
BiDense incorporates two techniques: the Distribution-adaptive Binarizer (DAB) and the Channel-adaptive Full-precision Bypass (CFB).
The DAB leverages the mean and mean absolute deviation (MAD) of full-precision inputs to dynamically determine the binarization thresholds and scaling factors, which captures the information from the distributions of real-valued variables, allowing it to adapt to unseen cases and preserve more messages.
Meanwhile, the CFB introduces minimal operations to construct full-precision bypasses for various channel size transformations, thereby minimizing real-valued information loss in binarized dense prediction networks.
By integrating these two techniques, BiDense outperforms all previous BNNs and closely approaches the performance of full-precision methods in dense prediction tasks of semantic segmentation and monocular depth estimation, with a minimal increase in parameters and operations.
Our contributions can be summarized as follows:
\begin{itemize}
    \item We propose BiDense, a generalized binarization framework for efficient dense prediction. To the best of our knowledge, this is the first work to apply binarization to the monocular depth estimation task. 
    \item We propose the Distribution-adaptive Binarizer (DAB), which adaptively binarizes activations according to the distribution of real-valued variable inputs, preserving full-precision information. 
    \item We propose the Channel-adaptive Full-precision Bypass (CFB), allowing full-precision skip connections for various channel size transformations in CNNs, thereby minimizing real-valued information loss.
    \item We evaluate BiDense in dense prediction tasks of semantic segmentation and monocular depth estimation, and show that our framework outperforms SOTA BNNs and achieves performance comparable to full-precision networks while requiring lower computational resources. 
\end{itemize}
\vspace{-1mm}
\section{Related Work}
\label{sec:related}

\vspace{-1mm}
\subsection{Dense Prediction}
\vspace{-1mm}
Dense prediction tasks involve pixel-level estimation in images, including semantic segmentation~\cite{zhao2017pyramid, chen2018encoder, xiao2018unified, cheng2021per, cheng2022masked, xie2021segformer}, monocular depth estimation~\cite{lee2019big, ranftl2021vision, yang2024depth, tu2022adabin, bhat2023zoedepth, huynh2020guiding}, keypoint detection~\cite{detone2018superpoint, yi2016lift, gleize2023silk, tyszkiewicz2020disk, zhao2022alike, revaud2019r2d2}, etc.
Neural networks designed for dense prediction typically consist of an encoder and a decoder. 
The encoder downsamples images to extract high-level semantic information, while the decoder upsamples features and integrates fine-grained details for the final prediction. 
Fully convolutional methods~\cite{long2015fully, chen2017rethinking, zhao2017pyramid, xiao2018unified, lee2019big, fu2019dual, tu2022adabin, huynh2020guiding} employ established CNN models~\cite{he2016deep, huang2017densely, liu2022convnet, xie2017aggregated} as backbones. In contrast, recent frameworks~\cite{ranftl2021vision, bhat2023zoedepth, jain2023oneformer, xie2021segformer, chen2022vision, cheng2022masked, strudel2021segmenter, yang2024depth} have begun to leverage advanced attention-based models~\cite{dosovitskiy2020image, bao2021beit, liu2021swin, liu2022swin, wang2023image, oquab2023dinov2} as encoders to further enhance accuracy.
While these methods have led to increasingly satisfactory performance, they often require extensive computational resources for precise estimations, which can be challenging for mobile devices. 
As a remedy, we propose model binarization for dense prediction, significantly reducing memory and computational costs while preserving accuracy to the greatest extent.

\vspace{-1mm}
\subsection{Binary Neural Networks}
\vspace{-1mm}
As the most extreme form of model quantization, model binarization requires neural networks to constrain both weights and activations to either $+1$ or $-1$~\cite{courbariaux2015binaryconnect, hubara2016binarized}.
This approach replaces floating-point operations with bitwise operations and full-precision parameters with 1-bit parameters, significantly accelerating inference and reducing memory usage. 
However, the binarization process eliminates full-precision information, leading to a substantial drop in accuracy. 
To address this issue, several methods~\cite{liu2020reactnet, qin2020forward, tu2022adabin, rastegari2016xnor, cai2024binarized, liu2022bit, he2023bivit} have been proposed to retain more real-valued information during binarization, thereby improving the performance of BNNs.
For instance, ReActNet~\cite{liu2020reactnet} shifts activations using learned parameters to achieve better binarization distributions;
% AdaBin~\cite{tu2022adabin} learns to select optimal binary sets;
BiT~\cite{liu2022bit} incorporates scaling factors for binary variables;
and BiSRNet~\cite{cai2024binarized} redistributes activations using learnable linear functions. 
Although these methods have shown improvements, they are not adaptive to unseen input distributions, which may result in suboptimal performance.
Furthermore, while previous BNNs have excelled in low-level vision tasks~\cite{cai2024binarized, zhangflexible, jiang2021training, xiabasic} and simple high-level tasks such as image classification and object detection~\cite{liu2020reactnet, tu2022adabin, liu2022bit, he2023bivit, qin2020forward}, only a few~\cite{zhou2022cellular, zhuang2019structured} have focused on the more resource-consuming dense prediction.
% Yet, their
In response, we propose BiDense to facilitate adaptive retention of real-valued information, enhancing the accuracy and generalization of dense prediction BNNs.

\begin{figure*}[htbp]
    \centering
    \includegraphics[width=0.9\linewidth]{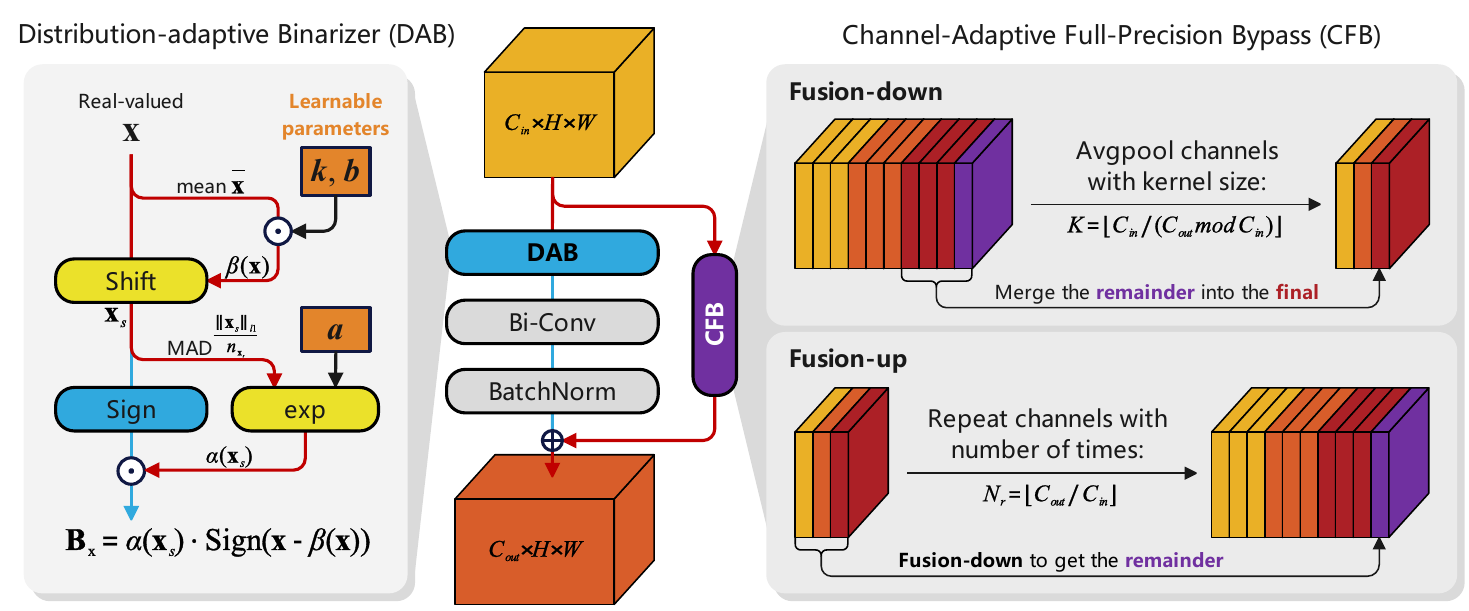}
    \vspace{-2mm}
    \caption{\textbf{Overview of a BiDense convolutional layer.} The BiDense layer integrates two techniques: \textbf{1)} The \textbf{Distribution-adaptive Binarizer (DAB)} adaptively binarizes activations based on input distributions, allowing it to retain more real-valued information. \textbf{2)} The \textbf{Channel-Adaptive Full-precision Bypass (CFB)} employs fusion-down and fusion-up to dynamically adjust the channel size of full-precision features, ensuring they align with the varying shapes of binary convolutional outputs while introducing minimal operations, which facilitates the propagation of real-valued and fine-grained information. \textbf{Bi-Conv} represents binary convolution. The \textbf{\textcolor[rgb]{0.75, 0, 0}{red}} arrow \textcolor[rgb]{0.75, 0, 0}{$\downarrow$} indicates the flow of full-precision information, while the \textbf{\textcolor[rgb]{0.19, 0.66, 0.87}{blue}} arrow \textcolor[rgb]{0.19, 0.66, 0.87}{$\downarrow$} denotes the flow of the binarized signal.}
    \label{fig:overview}
    \vspace{-4mm}
\end{figure*}

\vspace{-1mm}
\section{Methods}
\label{sec:methods}
\vspace{-1mm}
In this section, we first provide a brief overview of previous binarization methods.
Building on existing methods, we then introduce \textbf{BiDense}, a \textbf{Bi}nary neural network designed for \textbf{Dense} prediction tasks. 
BiDense incorporates two techniques: the \textit{Distribution-adaptive Binarizer} (DAB) and the \textit{Channel-adaptive Full-precision Bypass} (CFB), which aim to adaptively facilitate the retention and propagation of real-valued, fine-grained information, thereby enhancing performance in dense prediction tasks.
Finally, we will elaborate on the implementation details.
An overview of the BiDense architecture is summarized in \cref{fig:overview}.

\vspace{-1mm}
\subsection{Preliminaries}
\vspace{-1mm}
BNNs are required to constrain both the weights and the activations to either $+1$ or $-1$~\cite{hubara2016binarized, courbariaux2015binaryconnect}.
To achieve this, the sign function is typically used to binarize real-valued variables, resulting in the binarized values $\mathbf{B_x}$:
\begin{equation}
    \mathbf{B_x} = \text{Sign}(\mathbf{x}) = \left\{
        \begin{array}{rcl}
            +1 &  & \text{if $\mathbf{x}\geq0$} \\
            -1 &  & \text{otherwise},
        \end{array}
    \right.
    \label{eq:sign-function}
\end{equation}
where $\mathbf{x}$ denotes the full-precision parameters, which include both weights and activations.
% Due to the severe information loss in the binarization process, more advanced methods have been proposed to retain the information, improving the networks' performance.
% IR-Net~\cite{qin2020forward} balances the weights with zero-mean attribute by subtracting the mean to maximize the obtained binary weights' information entropy.
% ReActNet~\cite{liu2020reactnet} reshapes and shifts the activation distributions by the learnable threshold $\beta$, and the following BiT~\cite{liu2022bit} applies the scaling factor $\alpha$ to reduce the binarization error:
% \begin{equation}
%     x^b = \alpha \cdot \text{Sign}(x - \beta).
%     \label{eq:elastic-binarization-function}
% \end{equation}
According to \cite{qin2020forward}, the information entropy of $\mathbf{B_x}$ can be expressed as:
\begin{equation}
    \mathcal{H}(\mathbf{B_x}) = -p\ln{(p)}-(1-p)\ln{(1-p)},
\end{equation}
where $p$ is the probability of the value being $+1$, with $p\in(0, 1)$.
The entropy of the binary weights or activations serves as a measure of the information richness of the variables.
To retain more information, it is optimal for $p$ to equal 0.5 and maximize the entropy, resulting in an even distribution of the binarized values.
Ideally, this implies that the optimal binarization threshold corresponds to the median of the real-valued inputs.
However, obtaining the median through comparisons can be excessively resource-intensive for BNNs.
Thus, IR-Net~\cite{qin2020forward} proposes setting the threshold for the weights to the mean of the full-precision variables.
And ReActNet~\cite{liu2020reactnet} learns the thresholds $\beta$ as parameters for binarizing the activations:
\begin{equation}
    \mathbf{B_x} = \text{Sign}(\mathbf{x} - \beta).
    \label{eq:react-function}
\end{equation}

\begin{figure}
    \centering
    \footnotesize
    \tabcolsep=2pt
    \begin{tabular}{cccc}
        Real-valued & BNN~\cite{hubara2016binarized} & ReActNet~\cite{liu2020reactnet} & BiDense (Ours)
        \\
        \includegraphics[width=0.23\linewidth]{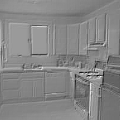} &
        \includegraphics[width=0.23\linewidth]{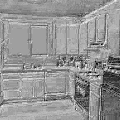} &
        \includegraphics[width=0.23\linewidth]{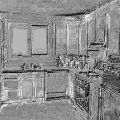} & 
        \includegraphics[width=0.23\linewidth]{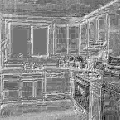} 
        \vspace{-0.5mm}
        \\
        $\mathcal{H}(\mathbf{B_x}) \rightarrow$ & 0.481 & 0.481 & \textbf{0.498}
        \vspace{0.5mm}
        \\
        \includegraphics[width=0.23\linewidth]{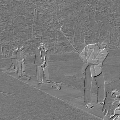} &
        \includegraphics[width=0.23\linewidth]{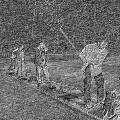} &
        \includegraphics[width=0.23\linewidth]{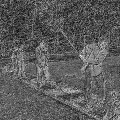} & 
        \includegraphics[width=0.23\linewidth]{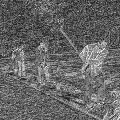} 
        \vspace{-0.5mm}
        \\
         $\mathcal{H}(\mathbf{B_x}) \rightarrow$ & 0.517 & 0.540 & \textbf{0.541}
        \vspace{0.5mm}
        \\
        \includegraphics[width=0.23\linewidth]{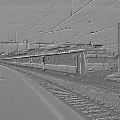} &
        \includegraphics[width=0.23\linewidth]{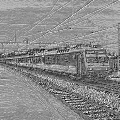} &
        \includegraphics[width=0.23\linewidth]{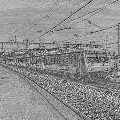} & 
        \includegraphics[width=0.23\linewidth]{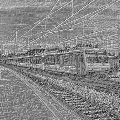} 
        \vspace{-0.5mm}
        \\
         $\mathcal{H}(\mathbf{B_x}) \rightarrow$ & 0.590 & 0.580 & \textbf{0.607}
    \end{tabular}
     \vspace{-2mm}
    \caption{\textbf{Average channel-wise information entropy} and \textbf{feature visualization} of activations before and after the first binarization in networks on the ADE20K~\cite{zhou2017scene} validation set. The feature visualization is represented by averaging the activations. 
    % BiDense effectively maximizes the information entropy through adaptive redistribution via the Distribution-adaptive Binarizer (DAB), preserving more real-valued messages.
    As indicated by information entropy, the fixed thresholds in BNN and ReActNet can result in suboptimal binarization and information loss, whereas BiDense retains more information by employing adaptive thresholds in the Distribution-adaptive Binarizer (DAB).
    }
    \vspace{-4mm}
    \label{fig:entropy}
\end{figure}

While learnable parameters can achieve optimal threshold values and maximize information entropy across the entire training set, their fixed nature after training may result in suboptimal solutions in localized contexts or when processing inputs from diverse distributions \cite{he2023bivit}.
As illustrated in \cref{fig:entropy}, fixed learned thresholds can lead to information loss when applied to unseen data, as indicated by a decrease in entropy.
To address the distribution variations and retain more information, we propose two techniques aimed at adaptively preserving real-valued signals based on the inputs, thereby facilitating more accurate dense predictions.

\vspace{-1mm}
\subsection{Distribution-adaptive Binarizer}
\vspace{-1mm}
Real-valued variables exhibit varying distributions across different image inputs.
As shown in \cref{fig:entropy}, binarizing activations with fixed thresholds can result in fluctuations in information richness after binarization when the input distributions vary.
To this end, we propose a dynamic approach for determining the binarization thresholds $\beta$, which adaptively redistributes and binarizes real-valued activations according to the input distributions:
\begin{align}
    \mathbf{B_x} &= \text{Sign}(\mathbf{x} - \beta(\mathbf{x})), 
    \label{eq:beta-DAB} \\
    \beta(\mathbf{x}) &= k \cdot \Bar{\mathbf{x}} + b,
    \label{eq:beta-formula}
\end{align}
where $k$, $b$ are learnable parameters, and $\Bar{\mathbf{x}}$ denotes the mean of the real-valued variables $\mathbf{x}$.
Inspired by \cite{qin2020forward}, we compute our thresholds using the mean to effectively capture the information from input distributions.
$b$ learns the average baseline across the entire dataset, akin to \cite{liu2020reactnet}.
Meanwhile, $k$ quantifies the influence rate of the current inputs, thereby balancing the threshold between global and localized distributions.
As indicated in \cref{fig:entropy} by information entropy, adaptively determining thresholds effectively enhances information richness after binarization in BiDense.

Following \cite{liu2022bit}, we incorporate a scaling factor $\alpha\in \mathbb{R}_+$ into the binarization function for activations.
% The scaling factor is designed to minimize the binarization error and preserve more real-valued information.
Bit~\cite{liu2022bit} learns the scale $\alpha$ and threshold $\beta$ as trainable parameters:
\begin{equation}
    \mathbf{B_x} = \alpha \cdot \text{Sign}(\mathbf{x} - \beta).
    \label{eq:elastic-binarization-function}
\end{equation}
% where $\alpha$ is initialized as $\alpha^*$.
Nonetheless, $\alpha$ in \cref{eq:elastic-binarization-function} remains fixed after training, preventing it from dynamically recovering the scaling information from the original full-precision values.
Therefore, we propose an approach to adaptively obtain $\alpha$ based on the original distributions of the real-valued data.
% Calculating $\alpha$ by minimizing the $l2$ error can be one of the solutions~\cite{liu2022bit}:
One potential solution involves calculating $\alpha$ by minimizing the $l_2$ error:
% which allows for the determination of optimal values for $\alpha$:
\begin{equation}
    \alpha^* = \mathop{\arg\min}\limits_{\alpha\in \mathbb{R}_+} \| \mathbf{x} - \alpha \mathbf{B_x} \|^2 = \frac{\| \mathbf{x} \|_{l1}}{n_\mathbf{x}},
    \label{eq:optimal-alpha}
\end{equation}
where $n_\mathbf{x}$ is the number of elements in $\mathbf{x}$.
Essentially, $\alpha^*$ preserves the mean absolute deviation (MAD) of the variables before and after binarization.
% While calculating $\alpha$ as $\alpha^*$ minimizes the binarization error, this solution does not necessarily enhance or impact the final performance in experiments.
% Although real-valued variables exhibit different distributions, most information is concentrated in the variations of 32 bits within each channel, while the MAD $\alpha^*$ as the scaling factors represents the scale information across different channels.
% Yet, the variation of MAD across channels is small~\cite{he2023bivit}, usually unable to provide 1-bit BNNs with sufficient information.
However, setting the scales of binarized activations to MAD may lead to significant numeric fluctuations in 1-bit variables.
Therefore, we introduce a learnable parameter that allows us to control the impact, and the full version of DAB is presented as:
\begin{align}
    \mathbf{x}_s &= \mathbf{x} - \beta(\mathbf{x}),
    \label{eq:shift-x} \\
    \mathbf{B_x} &= \alpha(\mathbf{x}_s) \cdot \text{Sign}(\mathbf{x}_s), 
    \label{eq:full-DAB} \\
    \alpha(\mathbf{x}_s) &= \exp{\left( a \cdot \left( \frac{\| \mathbf{x}_s \|_{l1}}{n_{\mathbf{x}_s}} - 1 \right) \right)},
    \label{eq:alpha-formula}
\end{align}
where $a$ is a layer-wise learnable parameter, $\mathbf{x}_s$ denotes variables shifted by the binarization threshold, and $n_{\mathbf{x}_s}$ is the number of elements in $\mathbf{x}_s$.
The parameter $\alpha$ in DAB is dynamically derived based on the MAD of the redistributed real-valued activations.
% We employ exponential function for several reasons.
The exponential function $\exp$ ensures that $\alpha$ remains greater than $0$, while also approximating a linear behavior as input $x$ approaches $1$:
\begin{equation}
    \lim_{x \to 1} \exp(a(x-1)) \approx ax.
    \label{eq:exp-linear}
\end{equation}
This linearity preserves the original proportional relationship of the scales.
As the influence rate of inputs, $a$ controls the suppression or amplification of the original scaling information.
When $a$ is initialized as $0$, $\alpha$ can be calculated with an initial value of $1$ during the training stage, thereby gradually improving performance from the baseline.
% Similar to \cite{he2023bivit, hu2018squeeze}, $\alpha(\mathbf{x}_s)$ can be interpreted as attention scores for the channels, rather than simply representing the scaling factor.
DAB binarizes real-valued variables adaptively based on the distribution characteristics of the mean and MAD, enabling enhanced information retention for unseen inputs.

\vspace{-1mm}
\subsection{Channel-adapative Full-precision Bypass}
\vspace{-1mm}
Neural networks for dense prediction tasks typically consist of an encoder and a decoder to enable pixel-level classification or regression.
Both stages incorporate multiple downsampling or upsampling modules, leading to feature channel size reductions or increases.
The transformations in channel sizes cause input and output dimension mismatches, obstructing full-precision skip connections, and leading to real-valued information loss and significant drops in prediction performance.
% , which are essential for preventing real-valued information loss during the binarization process. 
% This obstruction can lead to significant drops in prediction performance.
% Although \cite{cai2024binarized} introduces specific techniques to address the issue, it is limited to scenarios where the channel size is either halved or doubled.
% In contrast, most dense prediction networks~\cite{ranftl2021vision, xiao2018unified, zhao2017pyramid} incorporate convolutional operations with a wider range of channel transformations.
To facilitate full-precision bypassing across various channel size transformations for both downsampling and upsampling modules, we propose the Channel-adaptive Full-precision Bypass (CFB).

As shown in \cref{fig:overview}, we propose two straightforward techniques: fusion-down and fusion-up, which adaptively transform the channel size of full-precision features, retaining and aggregating real-valued information with minimal additional floating point operations.
For fusion-down, we apply 1D adaptive average pooling along the channel dimension to reduce the channel size.
Different from standard practice, for there is no spatial relationship across channels,  we set the kernel size and stride equal to avoid redundant calculations in overlaps.
The remainder input channels are merged into the final target channel.
In fusion-up, we utilize repetition to increase the channel sizes and employ the fusion-down operation to obtain the remainder target channels and integrate previous messages.
Specifically, given a full-precision input $\mathbf{x}_{in} \in \mathbb{R}^{C_{in} \times H \times W}$, where $C_{in}$, $H$, $W$ denote the input channel size, height, and width respectively, along with the target channel size $C_{out}$, the number of times to repeat each channel can be calculated as follows:
\begin{equation}
    N_r = \lfloor C_{out} / C_{in} \rfloor.
\end{equation}
After performing the repetition, the kernel size and stride for average pooling can be determined by:
\begin{equation}
    K = \lfloor C_{in} / (C_{out} \bmod C_{in}) \rfloor,
\end{equation}
and the last $K + (C_{in} \bmod K)$ channels are merged together for pooling.
The output of pooling operations is then concatenated with the results from the repetition, if applicable, allowing for adaptive transformations of channel size.
By performing the adaptive channel size transformations, the CFB facilitates the propagation of real-valued signals, retaining fine-grained, full-precision information.
Meanwhile, binary convolution focuses on learning the residuals, thereby maximizing the information utilization rate.

% \subsection{BiDense Block}
% Following 

\vspace{-1mm}
\subsection{Implementation Details}
\vspace{-1mm}
We build the BiDense convolutional layer based on the ReActNet~\cite{liu2020reactnet} convolutional layer, employing the same techniques for sign approximation functions and scaled binary weights.
As depicted in \cref{fig:overview}, Batch normalization is utilized within each BiDense layer to normalize the output of binary convolutions, effectively aligning the distributions of these outputs with the real-valued variables from the CFB.
To ensure minimal loss of real-valued information, we apply full-precision bypasses across all binarization modules.
% To validate the effectiveness of BiDense, we implement it on the convolutional neural network ConvNeXt-Tiny ~\cite{liu2022convnet}.
% To validate the effectiveness of BiDense, we implement it in two versions: one based on the convolutional neural network ConvNeXt~\cite{liu2022convnet} and another based on DPT~\cite{ranftl2021vision}, a hybrid architecture that combines a vision transformer (ViT)~\cite{dosovitskiy2020image} with a CNN. Our backbone for ConvNeXt is ConvNeXt-Tiny, while DPT is built on ViT-S.
% The prediction heads for different dense prediction tasks in BiDense are adapted from \cite{ranftl2021vision}. 
We leverage the AdamW~\cite{loshchilov2018decoupled} optimizer along with the OneCycle~\cite{smith2019super} learning rate scheduler to train our framework on a device with four RTX 3090 GPUs.

\vspace{-1mm}
\section{Experiments}
\label{sec:experiments}
\vspace{-1mm}
In this section, we first evaluate BiDense on two dense prediction tasks: semantic estimation and monocular depth estimation.
Then, we conduct ablation studies and present visualization results to further validate its effectiveness.

\vspace{0.5mm} \noindent \textbf{Baselines:}
For the full-precision 32-bit (FP32) baseline, we select ConvNeXt-Tiny~\cite{liu2022convnet} as the backbone encoder and UPerNet~\cite{xiao2018unified} as the decoder. 
The classification and regression head are derived from DPT~\cite{ranftl2021vision}. 
Building upon this FP32 baseline, we establish 1-bit baselines using previous advanced BNN methods, as well as BiDense with our proposed techniques.
% We select ConvNeXt-Tiny~\cite{liu2022convnet} as our full-precsion 32-bit (FP32) baseline, and various advanced BNNs as 1-bit baselines.
For a fair comparison, we provide results from two versions of FP32 baselines: one with pretrained weights on ImageNet~\cite{deng2009imagenet} for fine-tuning, referred to as ConvNeXt*, and another initialized with random weights for only training on the target datasets, referred to as ConvNeXt.
To adapt ConvNeXt for binarization, we modify the network by adding batch normalization after each convolutional layer to mitigate numeric instability caused by consecutive binary operations.
This adjustment may lead to a slight increase in parameters and computational operations.

\vspace{0.5mm} \noindent \textbf{Evaluation Protocols:}
For each task, we report the accuracy and error metrics following previous work to evaluate and compare the effectiveness.
We also provide the number of parameters (Params) and computational operations (OPs).
In line with \cite{hubara2016binarized}, the OPs for BNNs is caculated as $\text{OPs}^b = \text{OPs}^f / 64,$ where $\text{OPs}^f$ denotes the full-precision operations.
Similarly, the parameters of BNN is computed as $\text{Params}^b = \text{Params}^f / 32,$ where $\text{Params}^f$ indicates the full-precision parameters.
The total computational and memory costs of a model can be computed as $\text{OPs}=\text{OPs}^b+\text{OPs}^f$ and $\text{Params}=\text{Params}^b + \text{Params}^f$.

\vspace{-1mm}
\subsection{Semantic Segmentation}
\vspace{-1mm}
We evaluate BiDense using semantic segmentation, which is a pixel-level classification task.

\vspace{0.5mm} \noindent \textbf{Setup:}
Following \cite{ranftl2021vision}, we train and compare multiple methods on ADE20K~\cite{zhou2017scene} and PASCAL VOC 2012~\cite{everingham2010pascal} training and validation sets, as their test sets are not publicly available.
We preprocess the images following \cite{fu2019dual}, setting the input image size to $480 \times 480$ pixels for ADE20K and $500 \times 375$ pixels for PASCAL VOC during both training and inference.
The batch size is set to 32, and the maximum learning rate of the scheduler is configured to $1\times10^{-4}$.
We train the models for 50 epochs on ADE20K and 200 epochs on PASCAL VOC, utilizing the standard cross-entropy loss function.
Evaluation metrics, including pixel accuracy (pixAcc) and mean Intersection-over-Union (mIoU), are computed following the guidelines from \cite{ranftl2021vision}.
% We adopt the evaluation metrics from \cite{ranftl2021vision} to compute pixel accuracy (pixAcc) and mean Intersection-over Union (mIoU).

\begin{figure*}[htbp]
    \centering
    \footnotesize
    \tabcolsep=1.5pt
    \begin{tabular}{ccccccc}
        ConvNeXt~\cite{liu2022convnet} & BNN~\cite{hubara2016binarized} & ReActNet~\cite{liu2020reactnet} & AdaBin~\cite{tu2022adabin} & BiSRNet~\cite{cai2024binarized} & BiDense (Ours) & Ground Truth\\

        \includegraphics[width=0.135\linewidth]{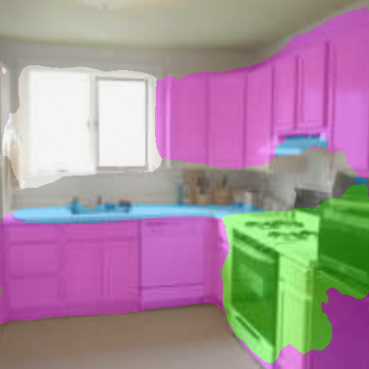} & 
        \includegraphics[width=0.135\linewidth]{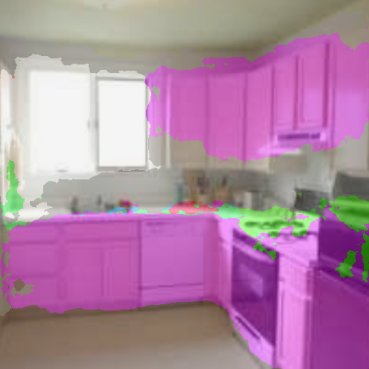} & 
        \includegraphics[width=0.135\linewidth]{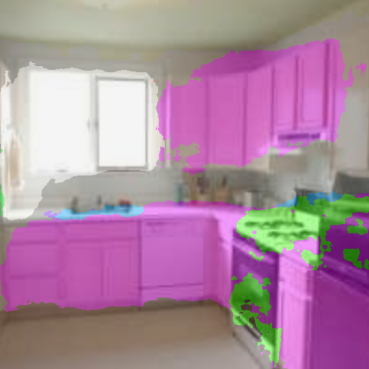} & 
        \includegraphics[width=0.135\linewidth]{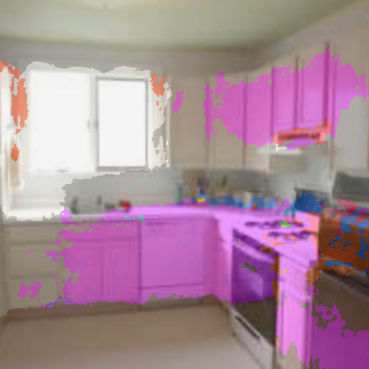} & 
        \includegraphics[width=0.135\linewidth]{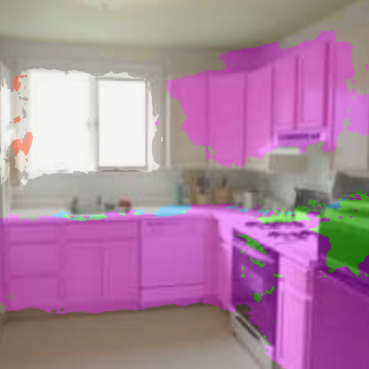} & 
        \includegraphics[width=0.135\linewidth]{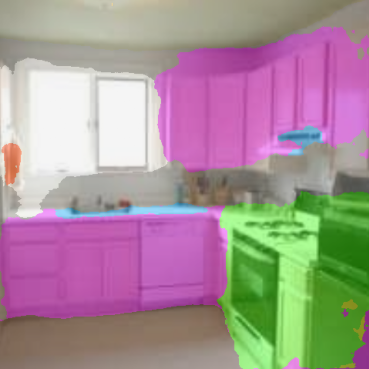} & 
        \includegraphics[width=0.135\linewidth]{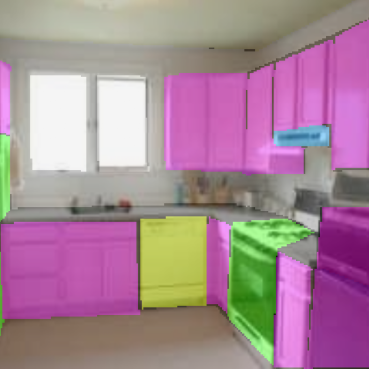} \\

        \includegraphics[width=0.135\linewidth]{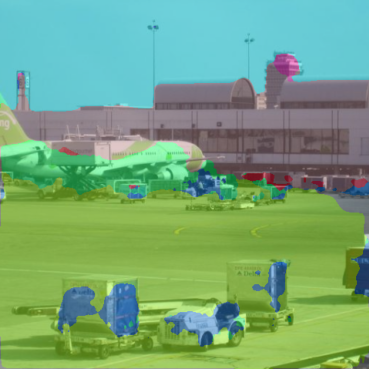} & 
        \includegraphics[width=0.135\linewidth]{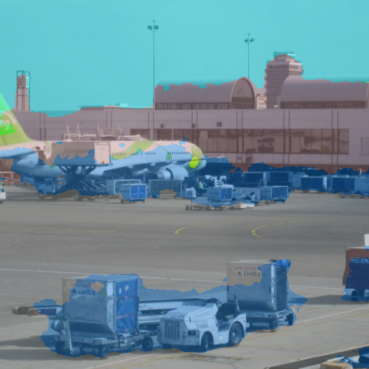} & 
        \includegraphics[width=0.135\linewidth]{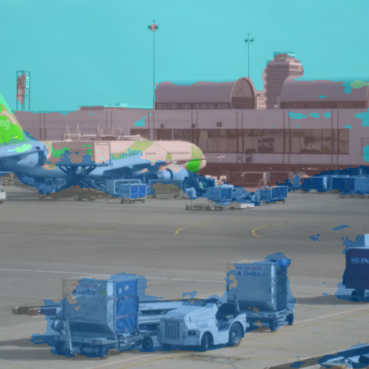} & 
        \includegraphics[width=0.135\linewidth]{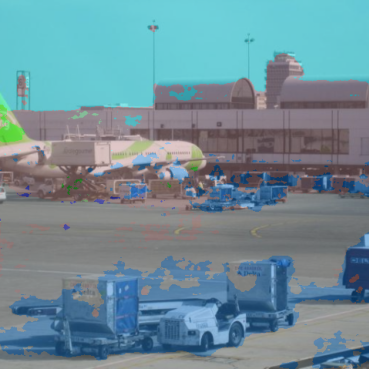} & 
        \includegraphics[width=0.135\linewidth]{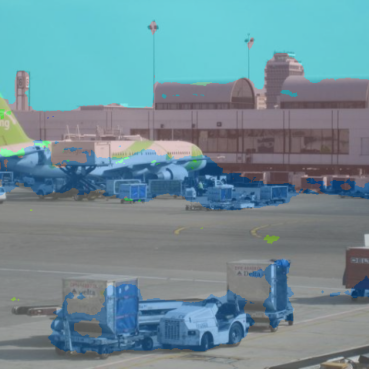} & 
        \includegraphics[width=0.135\linewidth]{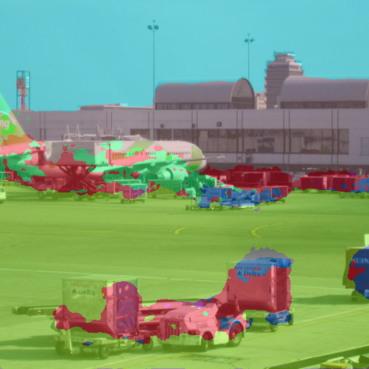} & 
        \includegraphics[width=0.135\linewidth]{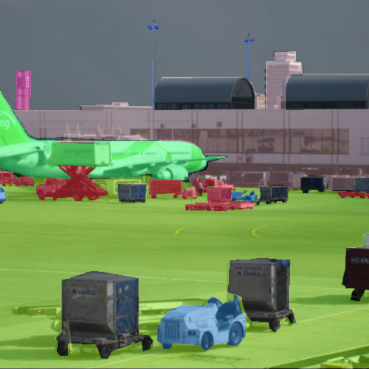} \\
        
    \end{tabular}
     \vspace{-2mm}
    \caption{\textbf{Qualitative Results of semantic segmentation on the ADE20K~\cite{zhou2017scene} validation set.} BiDense preserves full-precision semantic information, yielding more fine-grained and accurate segmentation than previous BNNs.}
    \label{fig:ade20k-viz}
    \vspace{-3mm}
\end{figure*}

\begin{table}
    \centering
    \footnotesize
    \tabcolsep=3.4pt
    \begin{tabular}{lccccc}
        \toprule[1.2pt]
        Method & Bit & Params (K) & OPs (G) & pixAcc (\%) & mIoU (\%) \\
        \midrule[1pt]
        % ConvNeXt*~\cite{liu2022convnet} & 32 & 40,023 & 127.43 & 79.48 & 40.96 \\
        % ConvNeXt~\cite{liu2022convnet} & 32 & 40,023 & 127.43 & 70.80 & 20.31 \\
        % \hline
        % BNN~\cite{hubara2016binarized} & 1 & 1,740 & 81.13 & 57.63 & 9.94 \\
        % ReActNet~\cite{liu2020reactnet} & 1 & 1,802 & 81.19 & 58.52 & 10.50 \\
        % AdaBin~\cite{} & 1 &  &  & 60.38 &  7.94 \\
        % BiSRNet~\cite{cai2024binarized} & 1 & 1,910 & 81.20 & 63.85 & 11.24 \\
        % BiDense-L (Ours) & 1 & 1,896 & 81.61 & \textbf{67.24} & \textbf{16.51} \\
        % \midrule[1pt]
        ConvNeXt*~\cite{liu2022convnet} & 32 & 40,020 & 288.36 & 79.48 & 40.96 \\
        ConvNeXt~\cite{liu2022convnet} & 32 & 40,023 & 288.37 & 70.81 & 20.39 \\
        \hline
        BNN~\cite{hubara2016binarized} & 1 & 1,470 & 4.84 & 61.69 & 8.68 \\
        ReActNet~\cite{liu2020reactnet} & 1 & 1,531 & 4.98 & 62.77 & 9.22 \\
        AdaBin~\cite{tu2022adabin} & 1 & 1,484 & 5.24 & 59.47 & 7.16 \\
        BiSRNet~\cite{cai2024binarized} & 1 & 1,639 & 5.07 & 62.85 & 9.74 \\
        BiDense (Ours) & 1 & 1,626 & 5.37 & \textbf{67.25} & \textbf{18.75} \\
        % DPT-ViT-S*~\cite{ranftl2021vision} & 32 & 35,314 &  & 80.41 & 44.46 \\
        % DPT-ViT-S~\cite{ranftl2021vision} & 32 & 35,314 &  & 69.09 & 20.67 \\
        % \hline
        % BiBert~\cite{qin2022bibert} & 1 & 2,161 &  & 49.33 & 4.61 \\
        % BiT~\cite{liu2022bit} & 1 & 2,193 &  & 59.53  & 9.26 \\
        % BiViT~\cite{he2023bivit} & 1 & 2,179 &  & 60.63  & 10.59 \\
        % BiDense-T (Ours) & 1 & 2,267 &  & \textbf{60.69} & \textbf{10.67} \\
        \bottomrule[1.2pt]
    \end{tabular}
    \vspace{-2mm}
    \caption{\textbf{Semantic segmentation results on the ADE20K~\cite{zhou2017scene} validation set.} BiDense outperforms other BNNs in accuracy while maintaining a limited increase in parameters and OPs, approaching the performance of FP32 baselines.}
    \label{tab:ade20k}
    \vspace{-4mm}
\end{table}

\begin{table}
    \centering
    \footnotesize
    \tabcolsep=2.1pt
    \begin{tabular}{lcccc}
        \toprule[1.2pt]
        Method & Bit & Topology & pixAcc (\%) & mIoU (\%) \\
        \midrule[1pt]
        % ConvNeXt*~\cite{liu2022convnet} & 32 & 40,023 & 127.43 & 79.48 & 40.96 \\
        % ConvNeXt~\cite{liu2022convnet} & 32 & 40,023 & 127.43 & 70.80 & 20.31 \\
        % \hline
        % BNN~\cite{hubara2016binarized} & 1 & 1,740 & 81.13 & 57.63 & 9.94 \\
        % ReActNet~\cite{liu2020reactnet} & 1 & 1,802 & 81.19 & 58.52 & 10.50 \\
        % AdaBin~\cite{} & 1 &  &  & 60.38 &  7.94 \\
        % BiSRNet~\cite{cai2024binarized} & 1 & 1,910 & 81.20 & 63.85 & 11.24 \\
        % BiDense-L (Ours) & 1 & 1,896 & 81.61 & \textbf{67.24} & \textbf{16.51} \\
        % \midrule[1pt]
        ConvNeXt*~\cite{liu2022convnet} & 32 & \multirow{2}{*}{ConvNeXt-Tiny, UPerNet} & 98.58 & 99.72 \\
        ConvNeXt~\cite{liu2022convnet} & 32 &  & 96.02 & 84.6 \\
        \hline
        % \hline
        LQ-Net~\cite{zhang2018lq} & 3 & \multirow{3}{*}{ResNet-18, FCN-32s} & - & 62.5 \\
        Group-Net~\cite{zhuang2019structured} & 1 &  & - & 65.1 \\
        CBNN~\cite{zhou2022cellular} & 1 &  & - & 66.5 \\
        \hline
        LQ-Net~\cite{zhang2018lq} & 3 & \multirow{2}{*}{ResNet-34, FCN-32s} & - & 70.4 \\
        Group-Net~\cite{zhuang2019structured} & 1 &  & - & 72.8 \\
        \hline
        LQ-Net~\cite{zhang2018lq} & 3 & \multirow{2}{*}{ResNet-50, FCN-32s} & - & 70.7 \\
        Group-Net~\cite{zhuang2019structured} & 1 &  & - & 71.0 \\
        \hline
        BNN~\cite{hubara2016binarized} & 1 & \multirow{5}{*}{ConvNeXt-Tiny, UPerNet} & 83.57 & 40.1 \\
        ReActNet~\cite{liu2020reactnet} & 1 &  & 83.38 & 42.1 \\
        AdaBin~\cite{tu2022adabin} & 1 &  & 78.27 & 25.2 \\
        BiSRNet~\cite{cai2024binarized} & 1 &  & 86.30 & 47.5 \\
        BiDense (Ours) & 1 &  & \textbf{92.73} & \textbf{75.2} \\
        % DPT-ViT-S*~\cite{ranftl2021vision} & 32 & 35,314 &  & 80.41 & 44.46 \\
        % DPT-ViT-S~\cite{ranftl2021vision} & 32 & 35,314 &  & 69.09 & 20.67 \\
        % \hline
        % BiBert~\cite{qin2022bibert} & 1 & 2,161 &  & 49.33 & 4.61 \\
        % BiT~\cite{liu2022bit} & 1 & 2,193 &  & 59.53  & 9.26 \\
        % BiViT~\cite{he2023bivit} & 1 & 2,179 &  & 60.63  & 10.59 \\
        % BiDense-T (Ours) & 1 & 2,267 &  & \textbf{60.69} & \textbf{10.67} \\
        \bottomrule[1.2pt]
    \end{tabular}
     \vspace{-2mm}
    \caption{\textbf{Semantic segmentation results on the PASCAL VOC 2012~\cite{everingham2010pascal} validation set.} BiDense outperforms prior quantized models, delivering finer-grained results.}
    \label{tab:pascal-voc}
    \vspace{-4mm}
\end{table}

\vspace{0.5mm} \noindent \textbf{Results:}
As shown in \cref{tab:ade20k}, BiDense outperforms other BNN baselines on ADE20K in terms of pixAcc and mIoU, while maintaining a comparable number of parameters and OPs
Compared to FP32 baselines, BiDense has $4.06\%$ of the parameters and $1.86\%$ of the OPs, with only $3.56\%$ and $1.64\%$ reductions of pixAcc and mIoU, respectively.
Moreover, when compared to previous BNNs, BiDense achieves remarkable improvements in terms of pixAcc and mIoU by $4.40\%$ and $9.01\%$, respectively, with only a limited increase in parameters and OPs.
\cref{tab:pascal-voc} further demonstrates the superior performance of BiDense on the PASCAL VOC datasets.
BiDense outperforms all prior quantized models and approaches the FP32 baseline in semantic segmentation, highlighting the effectiveness of our framework in preserving semantic content and real-valued information.
The qualitative results of semantic segmentation are also demonstrated in \cref{fig:ade20k-viz}, where BiDense displays highly fine-grained and accurate results compared to other baselines.

\vspace{-1mm}
\subsection{Monocular Depth Estimation}
\vspace{-1mm}
We also evaluate BiDense for monocular depth estimation, a pixel-level regression task in dense prediction.

\vspace{0.5mm} \noindent \textbf{Setup:}
Following \cite{ranftl2021vision}, we utilize the NYUv2~\cite{silberman2012indoor} and KITTI~\cite{geiger2012we} depth datasets to assess our method.
The images are preprocessed according to \cite{lee2019big}, with input image sizes of $416\times544$ for NYUv2 and $352\times704$ for KITTI, in both training and inference stages.
For NYUv2, we use a batch size of 32 and set the maximum learning rate of the scheduler to $6\times10^{-5}$.
For KITTI, the batch size is 64, and the maximum learning rate is $1\times10^{-4}$.
We employ the standard Silog loss function~\cite{eigen2014depth, lee2019big} to train all the methods.
Metrics are reported following the evaluation procedure outlined in \cite{ranftl2020towards}.
For $\delta_1$, $\delta_2$, and $\delta_3$, higher values indicate better performance, while for the error metrics AbsRel, RMSE, log10, and RMSE log, lower values are preferred.

\begin{figure*}[htbp]
    \centering
    \footnotesize
    \tabcolsep=1.5pt
    \begin{tabular}{ccccccc}
         & ConvNeXt~\cite{liu2022convnet} & BNN~\cite{hubara2016binarized} & ReActNet~\cite{liu2020reactnet} & BiSRNet~\cite{cai2024binarized} & BiDense (Ours) & Ground Truth \\
        \includegraphics[width=0.135\linewidth]{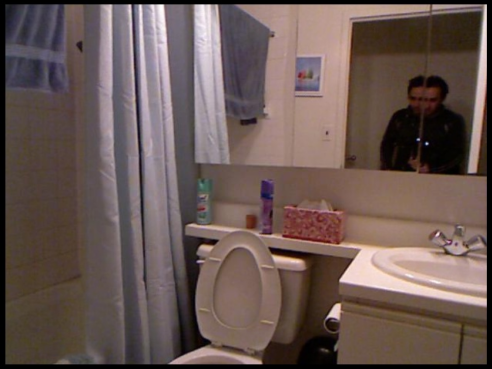} & 
        \includegraphics[width=0.135\linewidth]{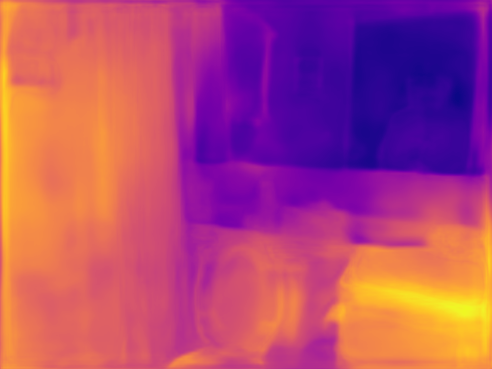} & 
        \includegraphics[width=0.135\linewidth]{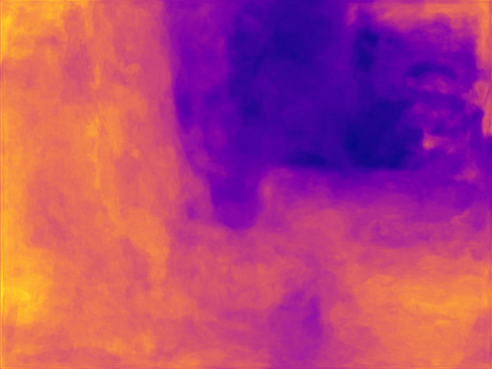} &
        \includegraphics[width=0.135\linewidth]{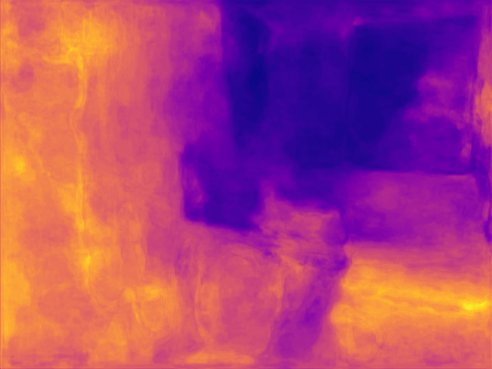} & 
        \includegraphics[width=0.135\linewidth]{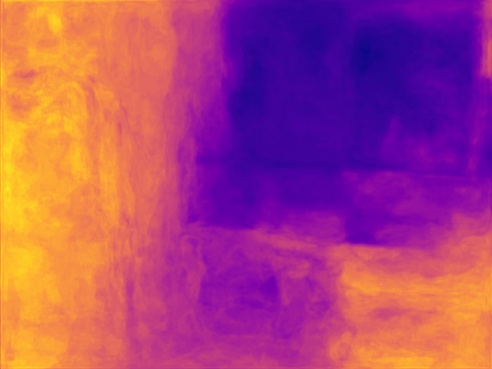} &
        \includegraphics[width=0.135\linewidth]{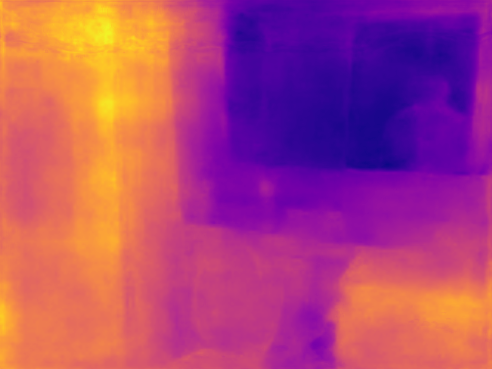} & 
        \includegraphics[width=0.135\linewidth]{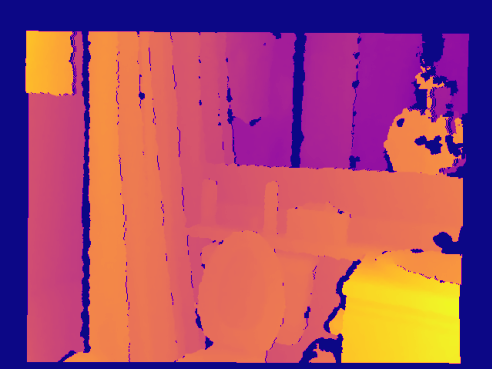}
        \\
        \includegraphics[width=0.135\linewidth]{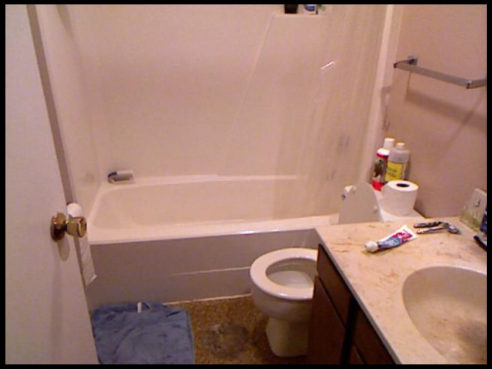} & 
        \includegraphics[width=0.135\linewidth]{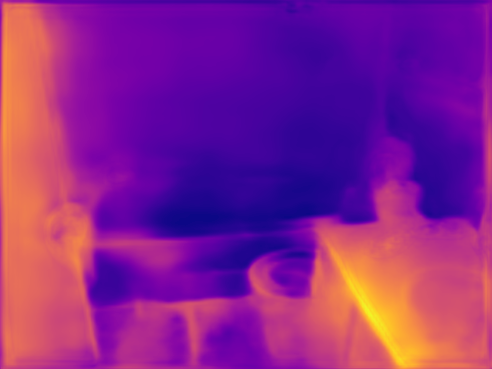} & 
        \includegraphics[width=0.135\linewidth]{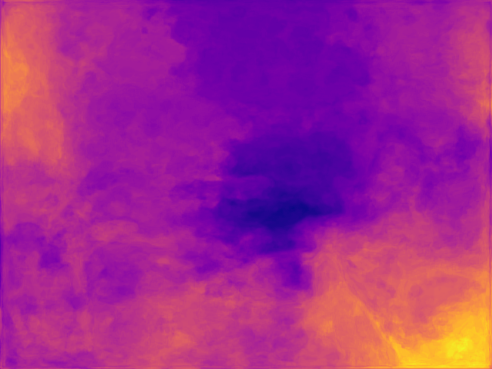} &
        \includegraphics[width=0.135\linewidth]{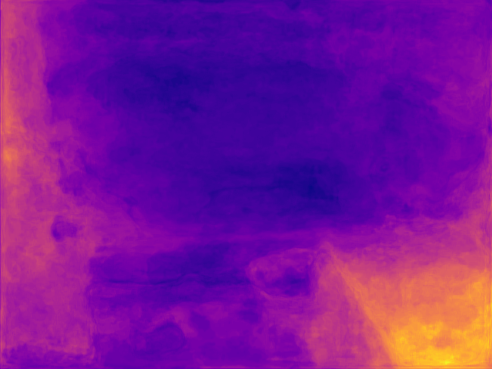} & 
        \includegraphics[width=0.135\linewidth]{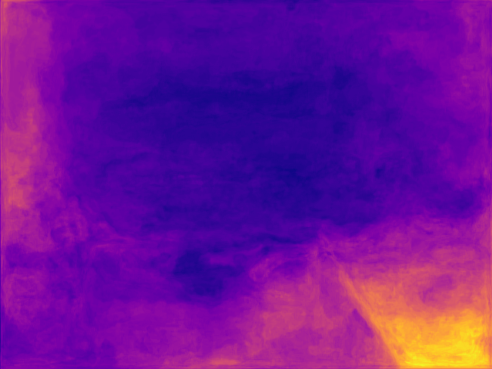} &
        \includegraphics[width=0.135\linewidth]{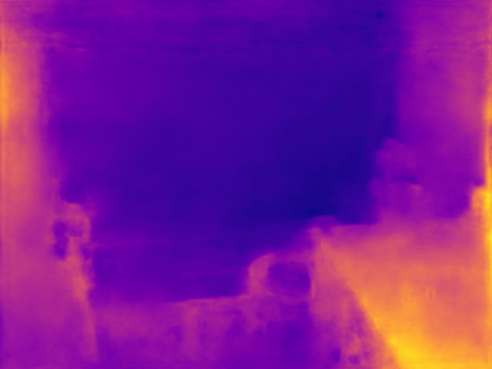} & 
        \includegraphics[width=0.135\linewidth]{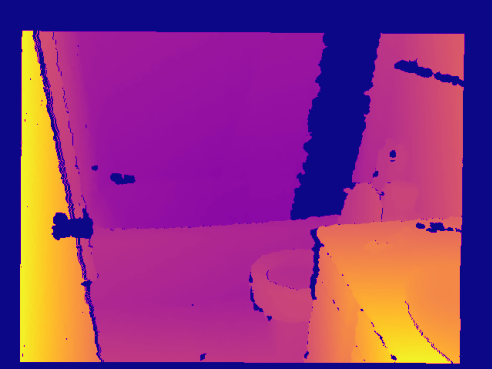}
    \end{tabular}

    \begin{tabular}{ccc}
         & ConvNeXt~\cite{liu2022convnet} & BiDense (Ours) \\
        \includegraphics[width=0.325\linewidth]{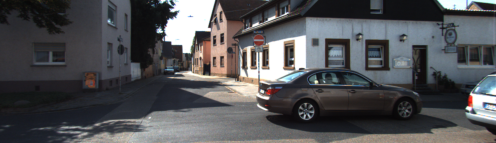} & 
        \includegraphics[width=0.325\linewidth]{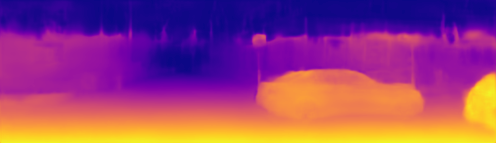} & 
        \includegraphics[width=0.325\linewidth]{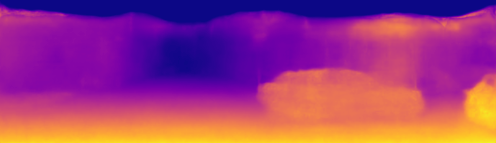}\\
        \includegraphics[width=0.325\linewidth]{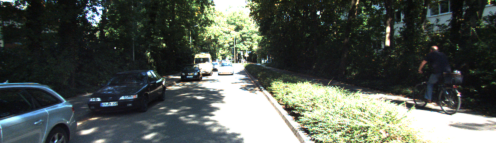} & 
        \includegraphics[width=0.325\linewidth]{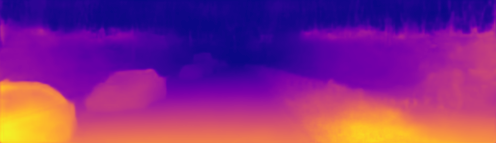} & 
        \includegraphics[width=0.325\linewidth]{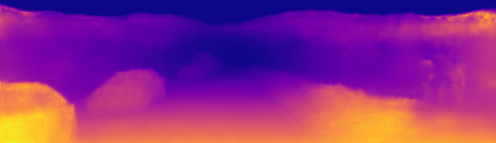}
    \end{tabular}
    \vspace{-2mm}
    \caption{\textbf{Qualitative results of monocular depth estimation results on the NYUv2~\cite{silberman2012indoor} and KITTI~\cite{geiger2012we} depth datasets.} BiDense retains real-valued information, resulting in more accurate performance and finer-grained depth predictions compared to previous BNNs.}
    \label{fig:depth-viz}
    \vspace{-1mm}
\end{figure*}

\begin{table*}
    \centering
    \footnotesize
    \tabcolsep=6.8pt
    \begin{tabular}{lc|cccccc|cccccc}
        \toprule[1.2pt]
        \multirow{2}{*}{Method} & \multirow{2}{*}{Bit} &  & \multicolumn{4}{c}{NYUv2~\cite{silberman2012indoor}} & & & \multicolumn{4}{c}{KITTI~\cite{geiger2012we}} & RMSE\\
         &  & $\delta_1$ & $\delta_2$ & $\delta_3$ & AbsRel & RMSE & log10 & $\delta_1$ & $\delta_2$ & $\delta_3$ & AbsRel & RMSE & log10 \\
        \midrule[1pt]
        ConvNeXt*~\cite{liu2022convnet} & 32 &  0.871&  0.979&  0.996&  0.117&  0.407&  0.050&  0.958&  0.996&  0.999&  0.065& 2.416& 0.094\\
        ConvNeXt~\cite{liu2022convnet} & 32 &  0.695&  0.916&  0.975&  0.201&  0.641&  0.081&  0.902&  0.980&  0.996&  0.092& 3.429& 0.137\\
        \hline
        BNN~\cite{hubara2016binarized} & 1 &  0.644&  0.887&  0.966&  0.223&  0.712&  0.091&  0.889&  0.976&  0.993&  0.102& 3.610& 0.150\\
        ReActNet~\cite{liu2020reactnet} & 1 &  0.664&  0.898&  0.969&  0.215&  0.686&  0.088&  0.893&  0.976&  0.993&  0.099& 3.449& 0.146\\
        AdaBin~\cite{tu2022adabin} & 1 & 0.537 & 0.826 & 0.945 & 0.256 & 0.851 & 0.112 &  0.853&  0.962& 0.990& 0.119& 4.016& 0.173\\
        BiSRNet~\cite{cai2024binarized} & 1 & 0.659 & 0.897 & 0.970 & 0.214 & 0.682 & 0.088 &  0.890&  0.976&  0.994&  0.101& 3.567& 0.147\\
        BiDense (Ours) & 1 & \textbf{0.712} & \textbf{0.921} & \textbf{0.978} & \textbf{0.189} & \textbf{0.610} & \textbf{0.078} & \textbf{0.896}&  \textbf{0.979}&  \textbf{0.995}&  \textbf{0.095}& \textbf{3.389}& \textbf{0.139}\\
        % \hline
        % BNN~\cite{hubara2016binarized} & &  &  &  &  &  &  &  &  &  &  &  & 1 & 1 \\
        % ReActNet~\cite{liu2020reactnet} & &  &  &  &  &  &  &  &  &  &  &  & 1 & 1 \\
        % AdaBin~\cite{tu2022adabin} & &  &  &  &  &  &  &  &  &  &  &  & 1 & 1 \\
        % BiSRNet~\cite{cai2024binarized} & &  &  &  &  &  &  &  &  &  &  &  & 1 & 1 \\
        % BiDense (Ours) & &  &  &  &  &  &  &  &  &  &  &  & 1 & 1 \\
        % \midrule[1pt]
        % DPT-ViT-S*~\cite{ranftl2021vision} & 32 & ViT-CNN &  &  & 0.960 & 0.996 & 0.999 & 0.066 & 2.335 & 0.093 \\
        % DPT-ViT-S~\cite{ranftl2021vision} & 32 & ViT-CNN &  &  &  &  &  &  &  & \\
        % \hline
        % BiBert~\cite{qin2022bibert} & 1 & ViT-CNN &  &  & 0.803 & 0.932 & 0.976 & 0.146 & 5.008 & 0.215 \\
        % BiT~\cite{liu2022bit} & 1 & ViT-CNN &  &  & 0.850 & 0.959 & 0.988 & 0.119 & 4.228 & 0.177 \\
        % BiViT~\cite{he2023bivit} & 1 & ViT-CNN &  &  & 0.857 & 0.962 & 0.989 & 0.118 & 4.106 & 0.173 \\
        % BiDense-T (Ours) & 1 & ViT-CNN &  &  &  &  &  &  &  &  \\
        \bottomrule[1.2pt]
    \end{tabular}
    \vspace{-2mm}
    \caption{\textbf{Monocular depth estimation results on the NYUv2~\cite{silberman2012indoor} and KITTI~\cite{geiger2012we} depth datasets.} Our framework outperforms all previous BNNs on both datasets. With the CFB, BiDense achieves even superior performance than the FP32 baseline on NYUv2.}
    \label{tab:depth}
    \vspace{-2mm}
\end{table*}

\vspace{0.5mm} \noindent \textbf{Results:}
\cref{tab:depth} presents the quantitative results of monocular depth estimation.
Again, BiDense outperforms all previous BNNs in the task.
On the KITTI dataset, our framework closely approaches the FP32 baseline, demonstrating comparable performance, while on the NYUv2 dataset, BiDense achieves even higher accuracy and lower errors across all metrics compared to the FP32 network.
We attribute this superior performance to the effectiveness of the CFB in retaining real-valued information and detailed content.
The skip connections and aggregation within the CFB minimize information loss,  allowing the binary convolutional layers to effectively capture deep features.
The qualitative results of depth estimation are illustrated in \cref{fig:depth-viz}, where BiDense delivers accurate depth predictions with competitive precision relative to the FP32 method.

\begin{figure*}[htbp]
    \centering
    \footnotesize
    \begin{tabular}{rr}
        \begin{subfigure}{0.45\linewidth}
            \centering
            \includegraphics[width=\linewidth]{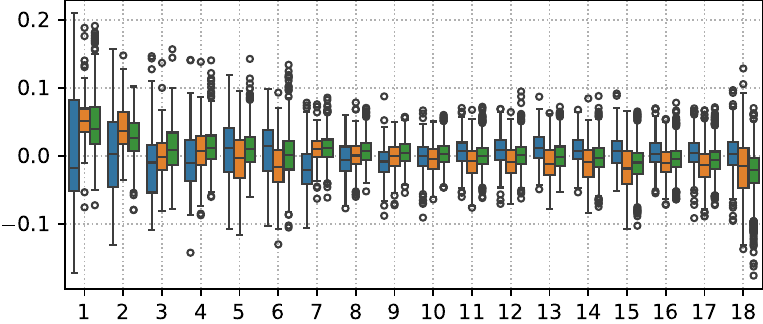}
            \caption{Distributions of $k$ across different blocks.}
            \label{fig:k-viz}
        \end{subfigure} & 
        \begin{subfigure}{0.45\linewidth}
            \centering
            \includegraphics[width=\linewidth]{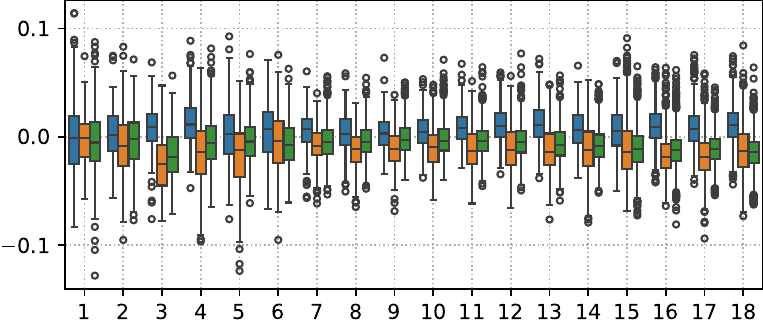}
            \caption{Distributions of $b$ across different blocks.}
            \label{fig:b-viz}
        \end{subfigure} 
        \\
        \begin{subfigure}{0.45\linewidth}
            \centering
            \includegraphics[width=\linewidth]{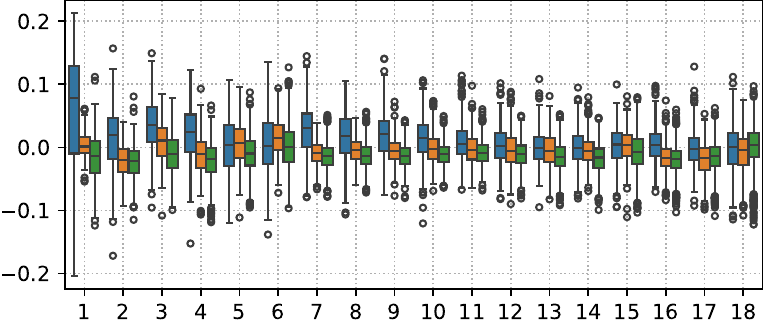}
            \caption{Distributions of $a$ across different blocks.}
            \label{fig:a-viz}
        \end{subfigure}
        &
        \begin{subfigure}{0.4362\linewidth}
            \centering
            \includegraphics[width=\linewidth]{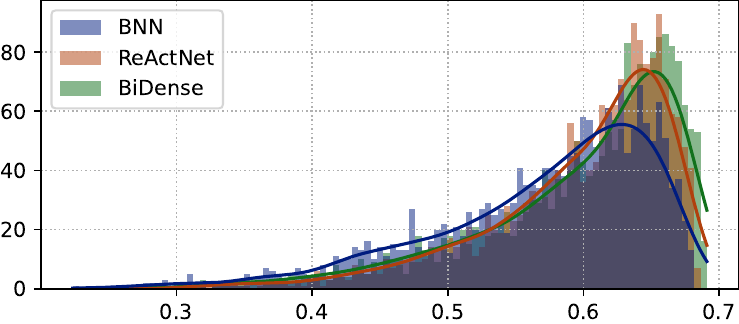}
            \caption{Information entropy after the first binarization process.}
            \label{fig:entropy-distribution}
        \end{subfigure}
    \end{tabular}
    \vspace{-3mm}
    \caption{\textbf{Statistics of DAB on the ADE20K~\cite{zhou2017scene} validation set.} \textbf{1)} \textbf{Panels (a), (b), and (c)} illustrate the statics of three learnable parameters, $k$, $b$, and $a$, from three different convolutional modules across the 18 ConvNeXt blocks, ranging from shallow to deep layers. The colors \textbf{\textcolor[rgb]{0.20, 0.45, 0.63}{blue}}, \textbf{\textcolor[rgb]{0.88, 0.51, 0.17}{orange}}, and \textbf{\textcolor[rgb]{0.23, 0.57, 0.23}{green}} represent the three consecutive convolutions within each block. Their distributions exhibit variability and all demonstrate bottleneck shapes throughout the layers. \textbf{2)} \textbf{Panel (d)} presents the statistics of average channel-wise information entropy after the initial binarization process in the networks for all inputs from the ADE20K validation set. BiDense generally shows higher entropy, indicating a greater average information richness after binarization and retains the most information.}
    \label{fig:visualization}
    \vspace{-5mm}
\end{figure*}

\begin{table}
    \centering
    \footnotesize
    \tabcolsep=3pt
    \begin{tabular}{lcccc}
        \toprule[1.2pt]
        Method & Params (K) & OPs(G) & pixAcc & mIoU \\
        \midrule[1pt]
        ConvNeXt~\cite{liu2022convnet} & 40,023 & 288.37 & 70.81 & 20.39 \\
        ReActNet~\cite{liu2020reactnet} & 1,531 & 4.98 & 62.77 & 9.22 \\
        \hline
        BiDense (Full) & 1,626 & 5.37 & \textbf{67.25} & \textbf{18.75} \\
        1) w/o DAB & 1,484 & 5.11 & 66.32 & 18.10 \\
        2) w/o CFB & 1,626 & 5.28 & 63.20 & 9.78 \\
        \hline
        3) w/o thresholds $\beta$ & 1,531 & 5.24 & 67.04 & 17.96 \\
        4) w/o $k\Bar{\mathbf{x}}$, $\beta(\mathbf{x})=b$ & 1,579 & 5.31 & 67.17 & 18.01 \\
        5) w/o $b$, $\beta(\mathbf{x})=k\Bar{\mathbf{x}}$ & 1,579 & 5.37 & 66.72 & 17.56 \\
        \hline
        6) w/o scaling factors $\alpha$ & 1,579 & 5.24 & 66.96 & 18.09 \\
        7) set $\alpha(\mathbf{x}_s)= \|\mathbf{x}_s\|_{l1}/n_{\mathbf{x}_s}$ & 1,579 &  5.37 & 49.3 & 4.97 \\
        8) set $\alpha$ as learnable parameters  & 1,626 & 5.31 & 65.14 & 15.10\\
        \midrule[1.2pt]
    \end{tabular}
    \vspace{-2mm}
    \caption{\textbf{Ablation study on the ADE20K~\cite{zhou2017scene} datasets.} 
    Our proposed techniques improve the performance of dense predictions.}
    \label{tab:ablation}
    \vspace{-4mm}
\end{table}

\vspace{-1mm}
\subsection{Ablation Study}
\vspace{-1mm}
We conduct ablation studies to evaluate the efficacy of the DAB and CFB using the ADE20K~\cite{zhou2017scene} datasets.
Specifically, we set BiDense as the baseline and compare it to various variants established by removing or substituting specific techniques with other methods.
\cref{tab:ablation} presents the detailed results of our ablation studies.
First, we demonstrate the effectiveness of the complete set of techniques:
\textbf{1)} \textbf{Removing the DAB} disables the adaptative binarization to different distributions of unseen inputs, which diminishes the information richness of features after binarization in BiDense and reduces estimation accuracy.
\textbf{2)} \textbf{Removing the CFB} results in a significant accuracy drop, as this eliminates the ability to retain real-valued information for convolutional layers undergoing various channel-size transformations in dense prediction networks, leading to considerable information loss.
Yet, with the DAB, BiDense still outperforms its 1-bit baseline, ReActNet.

\vspace{0.5mm} \noindent \textbf{Threshold:}
Then, we examine the roles of the adaptive thresholds $\beta(\mathbf{x})$ in our Distribution-adaptive Binarizer (DAB) by comparing the DAB to previous binarizers and its possible variants:
\textbf{3)} \textbf{Removing the thresholds $\beta$} prevents variable redistribution, which undermines information richness after binarization and results in accuracy drops.
\textbf{4)} \textbf{Setting $\beta$ as learnable parameters}, similar to \cite{liu2020reactnet}, fails to capture information from input distributions, leading to suboptimal binarization threshold values for unseen inputs.
\textbf{5)} \textbf{Without $b$}, setting $\beta(\mathbf{x}) = k\Bar{\mathbf{x}}$ causes fluctuations in binarization thresholds due to the mean $\Bar{\mathbf{x}}$ influenced by outliers, resulting in suboptimal values. 

\vspace{0.5mm} \noindent \textbf{Scale:}
Additionally, we compare the adaptive scaling factors $\alpha(\mathbf{x}_s)$ with other variants to study its role in performance.
\textbf{6)} \textbf{By removing the scaling factors $\alpha$}, the network struggles to retain essential real-valued information across different scales, resulting in a drop in accuracy.
\textbf{7)} \textbf{Setting the scales to MAD} introduces significant fluctuations as the variables transition from real-valued to 1-bit, complicating the training process and hindering robust performance.
\textbf{8)} \textbf{Making $\alpha$ learnable parameters}, similar to \cite{liu2022bit}, leads to accuracy drops on the validation set, because fixed parameters take on suboptimal values for unseen data.
The ablation study results validate the effectiveness of our proposed CFB and DAB.
By integrating these two techniques, BiDense enables adaptive real-valued information retention, leading to superior overall performance in dense predictions.

\vspace{-1mm}
\subsection{Visualization}
\vspace{-1mm}

In \cref{fig:visualization}, we visualize the statistics for BiDense, presenting the distributions of the learnable parameters and comparing the information entropy across different methods.

\vspace{0.5mm} \noindent \textbf{Learnable parameters:}
First, we report the value distributions of the three learnable parameters, $k$, $b$, and $a$, proposed in our DAB.
Given that the ConvNeXt backbone comprises three types of convolutions, each serving a distinct role within a ConvNeXt block, we visualize their parameter statistics separately using three different colors.
\cref{fig:k-viz}, \ref{fig:b-viz}, and \ref{fig:a-viz} illustrate the distributions of these parameters across the 18 blocks in the ConvNeXt backbone, spanning from shallow to deep layers.
The distributions exhibit variation across different convolutional modules within the blocks and across different network layers, suggesting that the parameters are optimized for improved binarization at various positions within the network.
The distributions of the three parameters across the blocks exhibit a bottleneck shape, mirroring the overall architecture of the backbone.

\vspace{0.5mm} \noindent \textbf{Information Entropy:}
\cref{fig:entropy-distribution} illustrates the information entropy of the activations after the first binarization process across different binary networks.
Overall, the entropy levels of BNN, ReActNet, BiDense increase in that order, indicating a greater richness of information as the methods evolve.
With the DAB, BiDense dynamically sets binarization thresholds to accommodate varying input value distributions, thereby enhancing information retention.

\vspace{-3mm}
\section{Conclusion}
\vspace{-1mm}
\label{sec:conclusion}
This paper proposes BiDense, a generalized binarization framework for efficient and accurate dense prediction tasks.
BiDense integrates two techniques: the Distribution-adaptive Binarizer (DAB) and the Channel-adaptive Full-Precision Bypass (CFB).
To adaptively retain more real-valued information for unseen data, the DAB utilizes the mean and the mean absolute deviation to derive binarization thresholds and scaling factors, which effectively capture real-valued information from input distributions.
Meanwhile, to facilitate full-precision bypasses when input and output feature dimension mismatch due to channel transformations in binary convolutional layers, the CFB incorporates minimal operations to adjust the channel size of full-precision features, propagating real-valued signals through various binary modules, thereby minimizing information loss.
We demonstrate that BiDense outperforms SOTA BNNs in the dense prediction tasks of semantic segmentation and monocular depth estimation, achieving accuracy comparable to that of full-precision networks while requiring significantly lower computational resources.
\vspace{-8mm}
{
    \small
    \bibliographystyle{ieeenat_fullname}
    \bibliography{main}
}

\clearpage
\appendix
\maketitlesupplementary

% \section{Rationale}
% \label{sec:rationale}
% % 
% Having the supplementary compiled together with the main paper means that:
% % 
% \begin{itemize}
% \item The supplementary can back-reference sections of the main paper, for example, we can refer to \cref{sec:intro};
% \item The main paper can forward reference sub-sections within the supplementary explicitly (e.g. referring to a particular experiment); 
% \item When submitted to arXiv, the supplementary will already included at the end of the paper.
% \end{itemize}
% % 
% To split the supplementary pages from the main paper, you can use \href{https://support.apple.com/en-ca/guide/preview/prvw11793/mac#:~:text=Delete%20a%20page%20from%20a,or%20choose%20Edit%20%3E%20Delete).}{Preview (on macOS)}, \href{https://www.adobe.com/acrobat/how-to/delete-pages-from-pdf.html#:~:text=Choose%20%E2%80%9CTools%E2%80%9D%20%3E%20%E2%80%9COrganize,or%20pages%20from%20the%20file.}{Adobe Acrobat} (on all OSs), as well as \href{https://superuser.com/questions/517986/is-it-possible-to-delete-some-pages-of-a-pdf-document}{command line tools}.

% \section{Additional Results}
\vspace{-2mm}
\section{DPT Architecture}
\vspace{-1mm}
In addition to CNNs, recent methods have utilized transformer architectures and transformer-CNN hybrid architectures for dense predictions.
To further validate the effectiveness of BiDense, we apply and compare various binary methods on DPT~\cite{chen2022vision}, a hybrid dense prediction network consisting of a transformer encoder and a CNN decoder.
\vspace{0.5mm} \noindent \textbf{Setup:}
We select ViT-S~\cite{dosovitskiy2020image} as the backbone encoder and present the results for two versions of the FP32 baseline: one initialized with DINOv2~\cite{oquab2023dinov2} weights, referred to as DPT*, and another initialized with random weights, referred to as DPT.
For the transformer component, we leverage existing transformer binarization methods, including BiBert~\cite{qinbibert}, BiT~\cite{liu2022bit}, and BiViT~\cite{he2023bivit}, to establish our BNN baselines.
In the CNN component, we use ReActNet~\cite{liu2020reactnet} as the binary decoder to incorporate with the three binary transformer encoders.
We construct BiDense-T by employing BiViT as the transformer encoder and applying our proposed techniques to the CNN decoder.
We refer to the fully convolutional network based on ConvNeXt as BiDense-C, which is primarily discussed in the main text.

\begin{table}
    \centering
    \footnotesize
    \tabcolsep=3pt
    \begin{tabular}{lccccc}
        \toprule[1.2pt]
        Method & Bit & Params (K) & OPs (G) & pixAcc (\%) & mIoU (\%) \\
        \midrule[1pt]
        ConvNeXt*~\cite{liu2022convnet} & 32 & 40,020 & 288.36 & 79.48 & 40.96 \\
        ConvNeXt~\cite{liu2022convnet} & 32 & 40,023 & 288.37 & 70.81 & 20.39 \\
        \hline
        BNN~\cite{hubara2016binarized} & 1 & 1,470 & 4.84 & 61.69 & 8.68 \\
        ReActNet~\cite{liu2020reactnet} & 1 & 1,531 & 4.98 & 62.77 & 9.22 \\
        % AdaBin~\cite{tu2022adabin} & 1 & 1,484 & 5.24 & 59.47 & 7.16 \\
        BiSRNet~\cite{cai2024binarized} & 1 & 1,639 & 5.07 & 62.85 & 9.74 \\
        BiDense-C (Ours) & 1 & 1,626 & 5.37 & \textbf{67.25} & \textbf{18.75} \\
        \midrule[1pt]
        DPT*~\cite{ranftl2021vision} & 32 & 35,313  & 232.20 & 78.87 & 37.73 \\
        DPT~\cite{ranftl2021vision} & 32 & 35,313 & 232.20 & 67.17 & 18.04 \\
        \hline
        BiBert~\cite{qin2022bibert} & 1 & 1,955 & 4.14 & 50.83 &	4.81 \\
        BiT~\cite{liu2022bit} & 1 & 1,988 & 4.31 & 59.12  & 8.66 \\
        BiViT~\cite{he2023bivit} & 1 & 1,974 & 4.33 & 60.09  & 9.40 \\
        BiDense-T (Ours) & 1 & 1,997 & 4.55 & \textbf{62.43} & \textbf{14.27} \\
        \bottomrule[1.2pt]
    \end{tabular}
    \vspace{-2mm}
    \caption{Semantic segmentation results of the DPT architecture on ADE20K~\cite{zhou2017scene} val, expanding upon Tab. 1 in the main text.
    }
    \label{tab:dpt-ade20k}
    \vspace{-4mm}
\end{table}

\begin{figure}[htbp]
    \centering
    \footnotesize
    \tabcolsep=1.5pt
    \begin{tabular}{cccc}
        & ConvNeXt~\cite{liu2022convnet} & BiDense (Ours) & Ground Truth \\
        \includegraphics[width=0.235\linewidth]{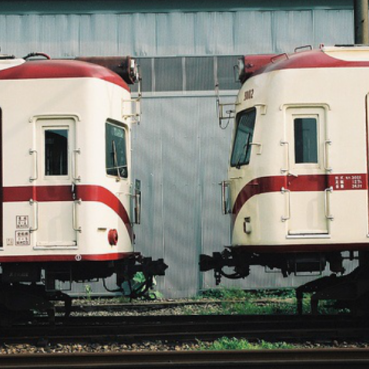} & 
        \includegraphics[width=0.235\linewidth]{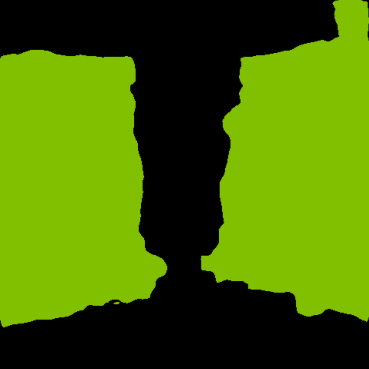} & 
        \includegraphics[width=0.235\linewidth]{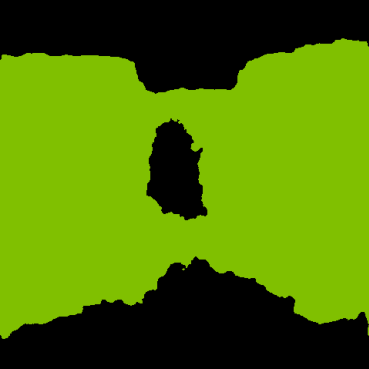} &
        \includegraphics[width=0.235\linewidth]{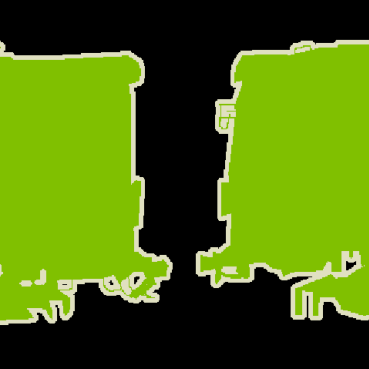} \\

        \includegraphics[width=0.235\linewidth]{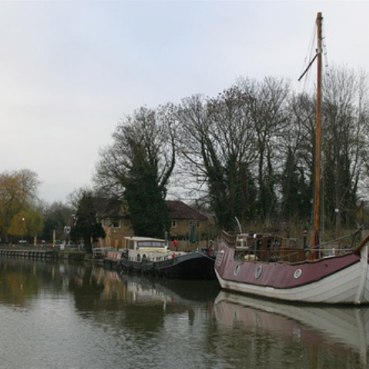} & 
        \includegraphics[width=0.235\linewidth]{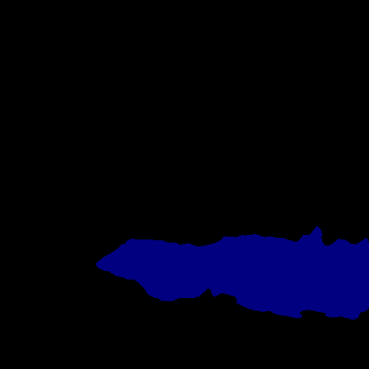} & 
        \includegraphics[width=0.235\linewidth]{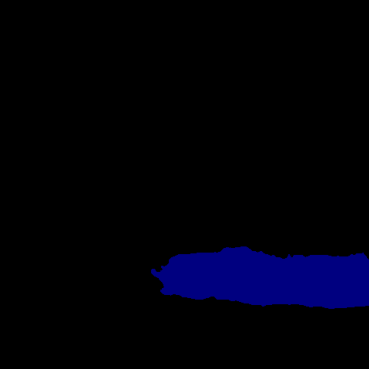} &
        \includegraphics[width=0.235\linewidth]{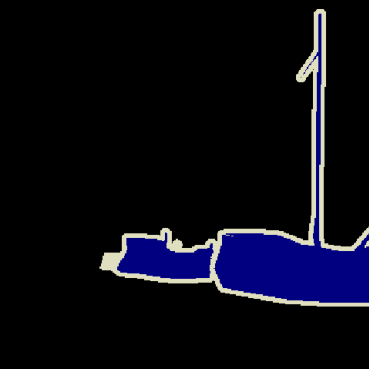} \\

        \includegraphics[width=0.235\linewidth]{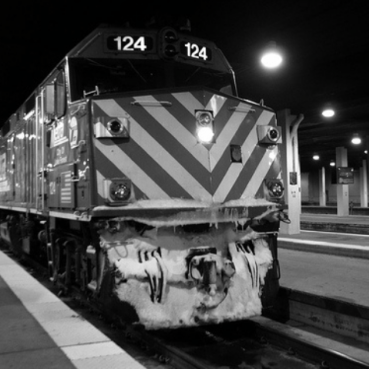} & 
        \includegraphics[width=0.235\linewidth]{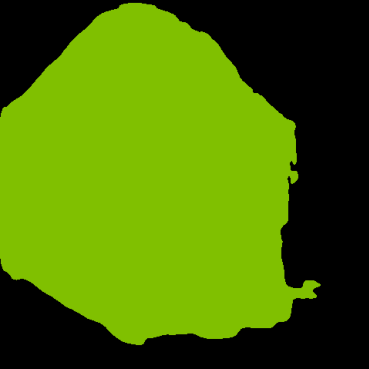} & 
        \includegraphics[width=0.235\linewidth]{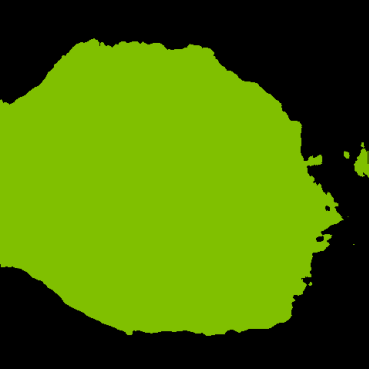} &
        \includegraphics[width=0.235\linewidth]{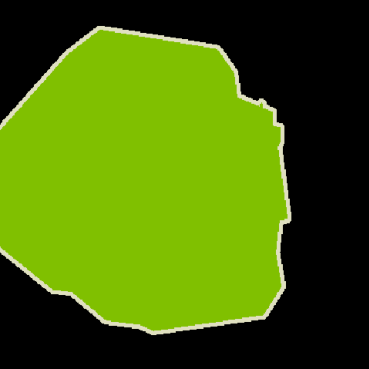} \\

        \includegraphics[width=0.235\linewidth]{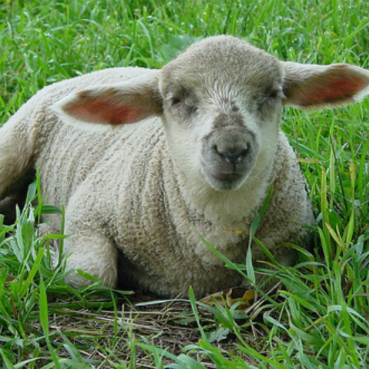} & 
        \includegraphics[width=0.235\linewidth]{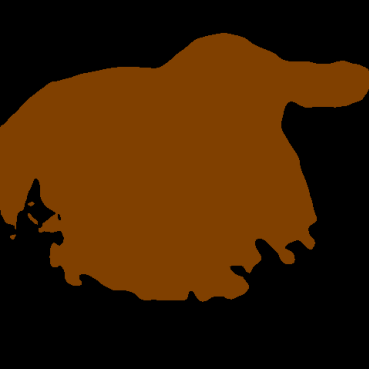} & 
        \includegraphics[width=0.235\linewidth]{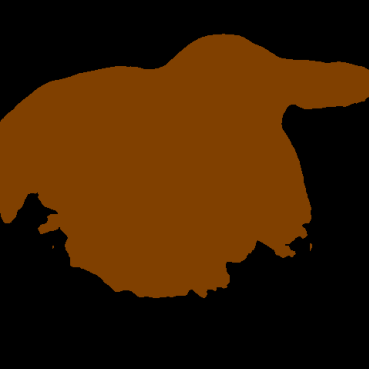} &
        \includegraphics[width=0.235\linewidth]{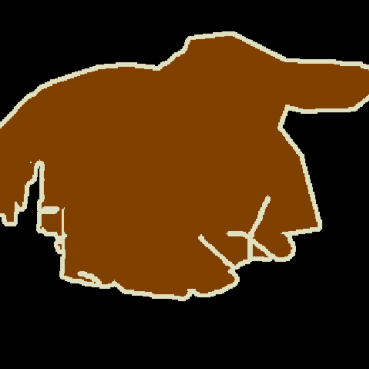} \\

        \includegraphics[width=0.235\linewidth]{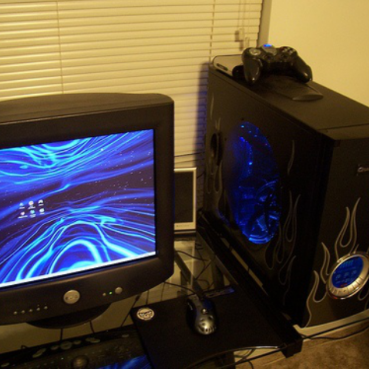} & 
        \includegraphics[width=0.235\linewidth]{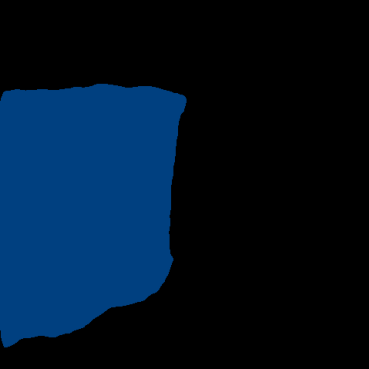} & 
        \includegraphics[width=0.235\linewidth]{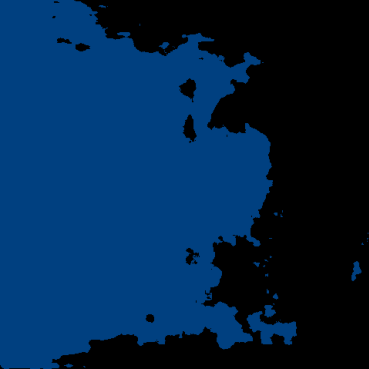} &
        \includegraphics[width=0.235\linewidth]{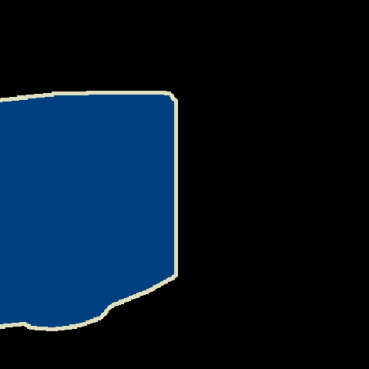} \\
        
        \includegraphics[width=0.235\linewidth]{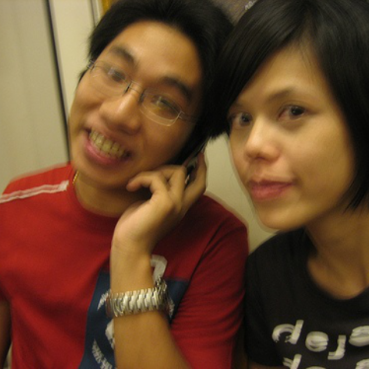} & 
        \includegraphics[width=0.235\linewidth]{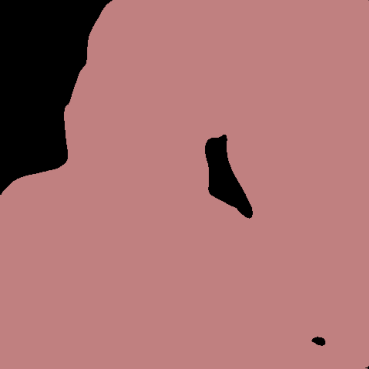} & 
        \includegraphics[width=0.235\linewidth]{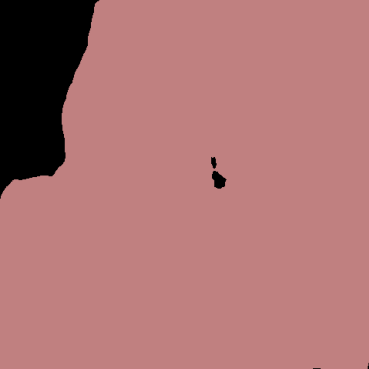} &
        \includegraphics[width=0.235\linewidth]{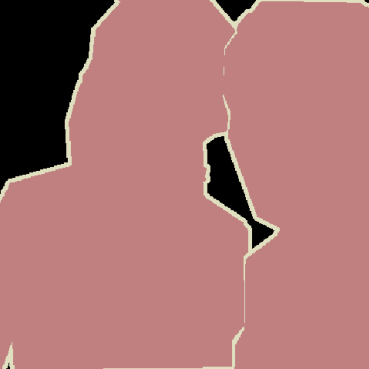} \\
        
    \end{tabular}
    \vspace{-2mm}
    \caption{Qualitative semantic segmentation results on the PASCAL VOC 2012~\cite{everingham2010pascal} validate set.}
    \vspace{-4mm}
    \label{fig:add-pascal}
\end{figure}

\vspace{0.5mm} \noindent \textbf{Results:}
\cref{tab:dpt-ade20k} demonstrates that BiDense-T outperforms other binary transformer baselines.
In semantic segmentation, our framework achieves 5.66\% of the parameters and 1.96\% of the operations (OPs) compared to FP32 DPT, while approaching its performance.

% \section{Kernel Size}
\vspace{-2mm}
\section{AdaBin}
\vspace{-1mm}

We observe that AdaBin~\cite{tu2022adabin} exhibits inferior performance across various tasks. 
Unlike other BNN baselines, the key distinction of AdaBin is that it does not perform redistribution for variables before non-linear layers. 
To ensure a fair comparison, we adhere strictly to the official implementation of AdaBin CNN blocks. 
However, this approach may not be optimal for dense prediction tasks and could result in a decrease in accuracy.

\vspace{-2mm}
\section{Limitations}
\vspace{-1mm}
% BiDense techniques are not suitable for self-attention modules or transformer-based networks.
% Previous work~\cite{liu2022bit, he2023bivit} and our experiments have shown that applying advanced methods for activations in the linear layers in self-attention modules may lead to significant overfitting, severely impairing the inference accuracy, while more sophisticated weight binarization techniques may be harmless.
% This blocks the application of DAB in transformers.
% Nonetheless, for most fully convolutional or transformer-CNN hybrid networks for dense predictions, BiDense remains superior to previous BNNs.

BiDense techniques are not well-suited for self-attention modules or transformer-based networks. Previous studies~\cite{liu2022bit, he2023bivit} and our experiments have shown that applying advanced binarization methods for activations in the linear layers of self-attention modules can lead to significant overfitting, severely impairing inference accuracy. In contrast, more sophisticated weight binarization techniques tend to have a positive impact. This limitation hinders the application of DAB in transformers. Nevertheless, for most fully convolutional networks or transformer-CNN hybrid architectures used for dense predictions, BiDense remains superior to previous BNN approaches.

% \vspace{-2mm}
% \section{Qualitative Results}
% \vspace{-1mm}

% We show more qualitative results of dense predictions in \cref{fig:add-pascal}, 
\begin{figure*}
    \centering
    \footnotesize
    \tabcolsep=1.5pt
    \begin{tabular}{ccccccc}
         & ConvNeXt~\cite{liu2022convnet} & BNN~\cite{hubara2016binarized} & ReActNet~\cite{liu2020reactnet} & BiSRNet~\cite{cai2024binarized} & BiDense (Ours) & Ground Truth \\
        \includegraphics[width=0.135\linewidth]{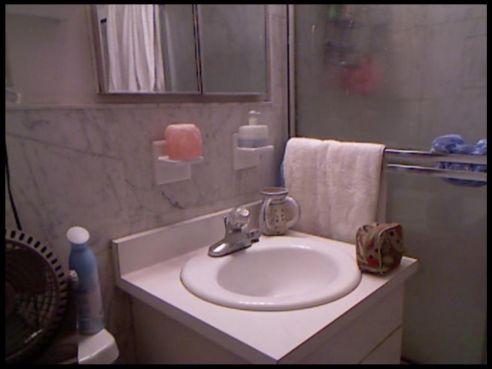} & 
        \includegraphics[width=0.135\linewidth]{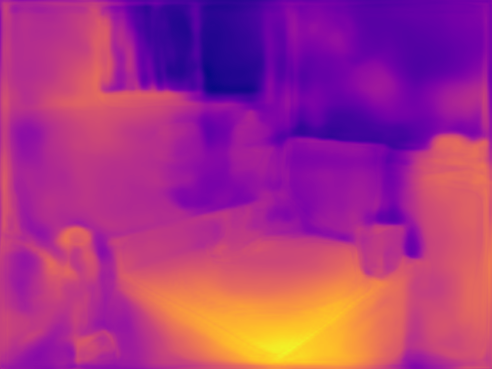} & 
        \includegraphics[width=0.135\linewidth]{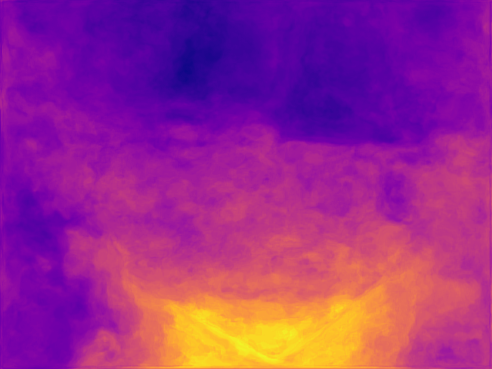} &
        \includegraphics[width=0.135\linewidth]{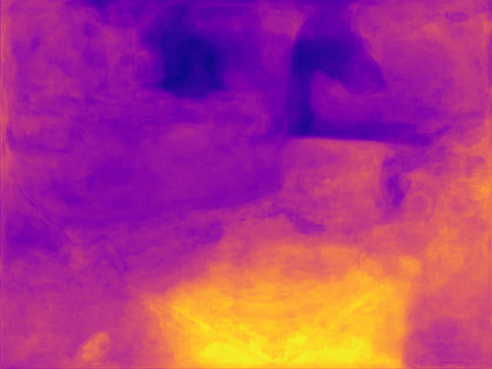} & 
        \includegraphics[width=0.135\linewidth]{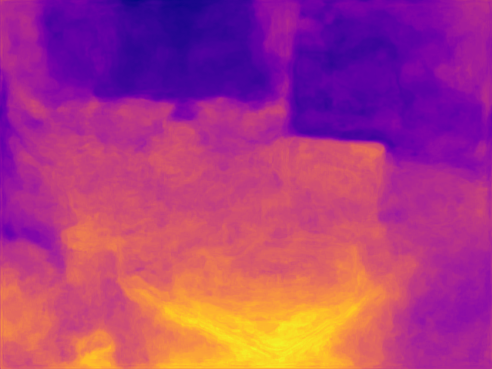} &
        \includegraphics[width=0.135\linewidth]{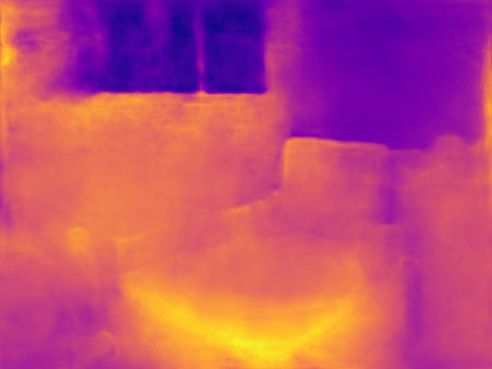} & 
        \includegraphics[width=0.135\linewidth]{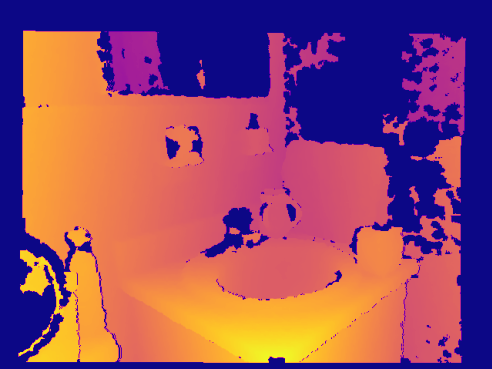}
        \\
        \includegraphics[width=0.135\linewidth]{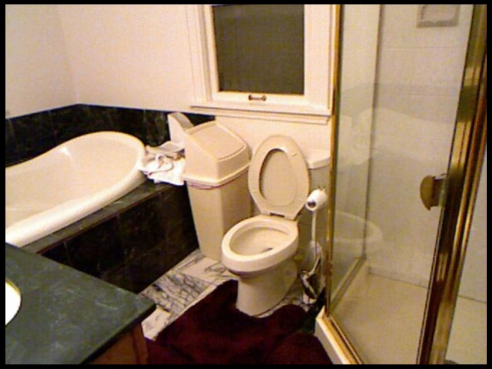} & 
        \includegraphics[width=0.135\linewidth]{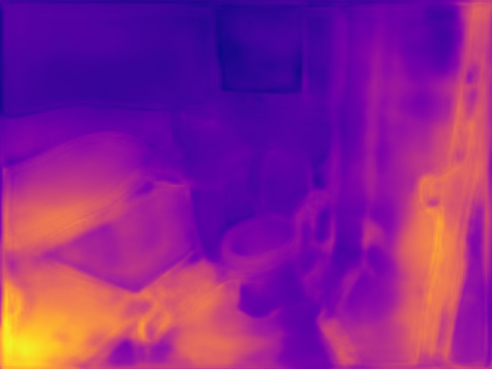} & 
        \includegraphics[width=0.135\linewidth]{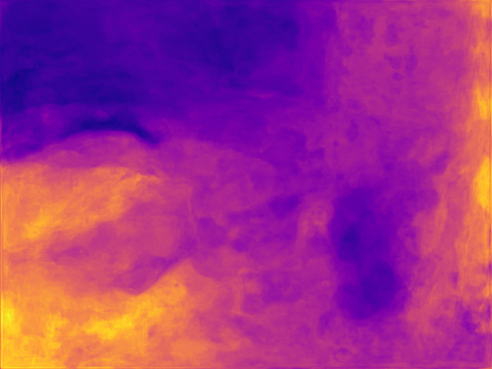} &
        \includegraphics[width=0.135\linewidth]{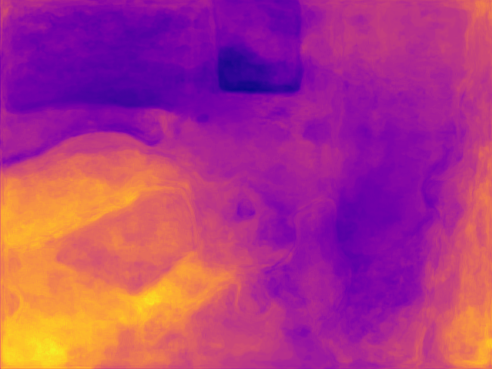} & 
        \includegraphics[width=0.135\linewidth]{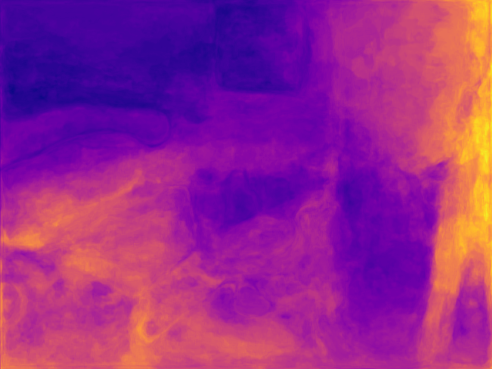} &
        \includegraphics[width=0.135\linewidth]{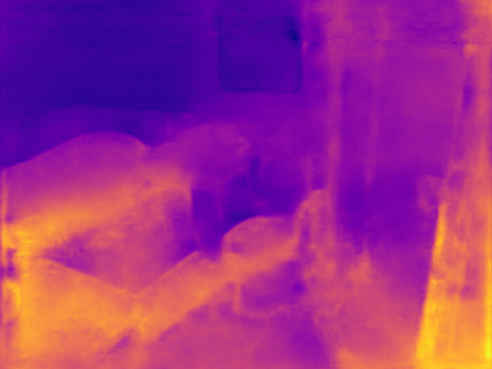} & 
        \includegraphics[width=0.135\linewidth]{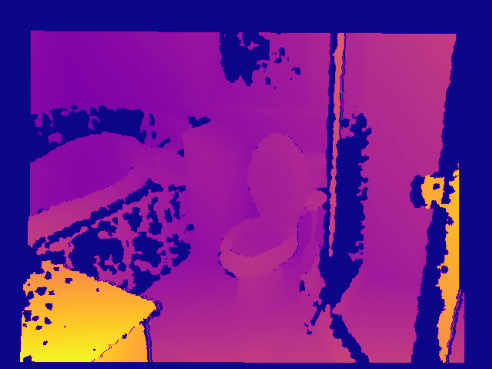}
        \\
        \includegraphics[width=0.135\linewidth]{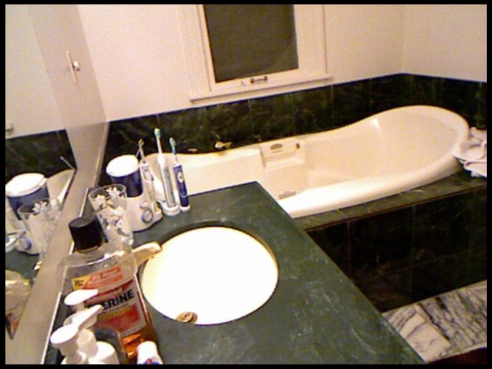} & 
        \includegraphics[width=0.135\linewidth]{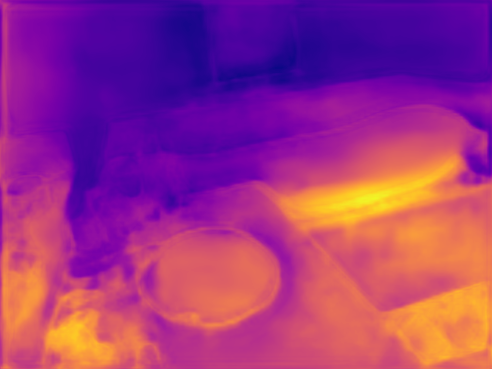} & 
        \includegraphics[width=0.135\linewidth]{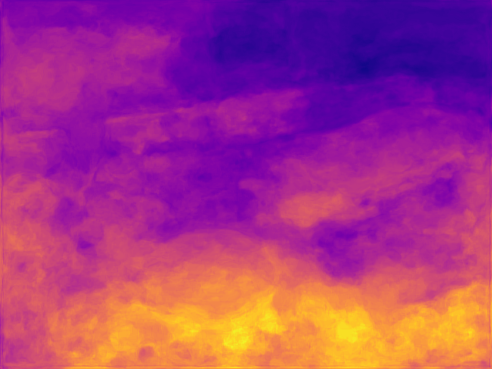} &
        \includegraphics[width=0.135\linewidth]{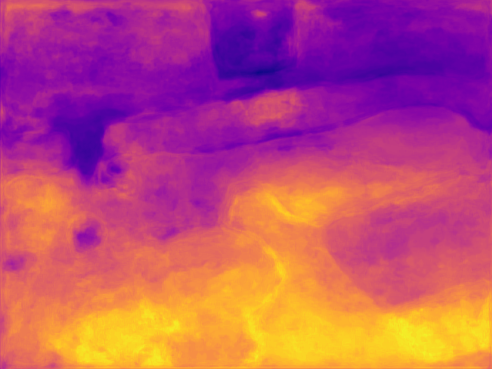} & 
        \includegraphics[width=0.135\linewidth]{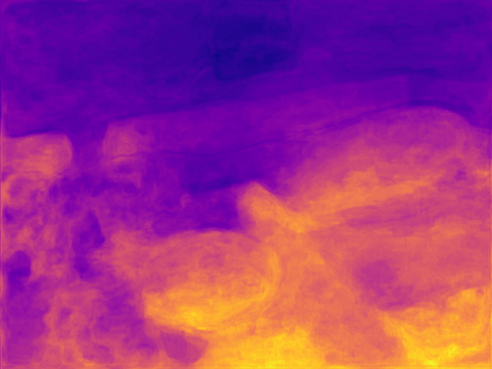} &
        \includegraphics[width=0.135\linewidth]{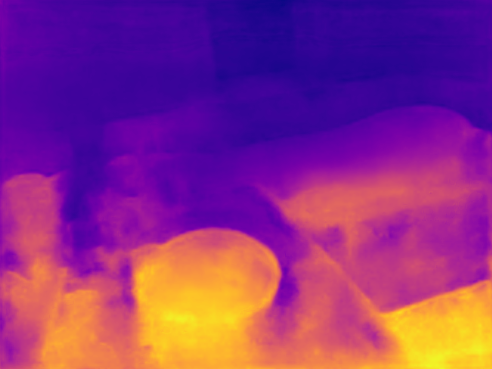} & 
        \includegraphics[width=0.135\linewidth]{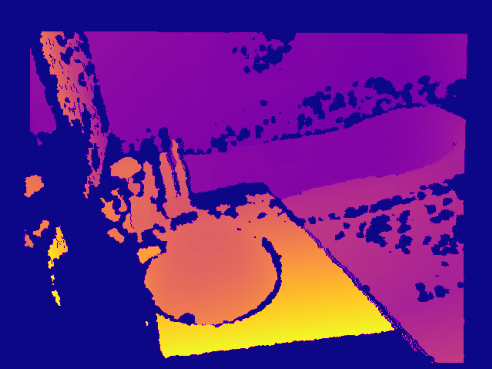}
        \\
        \includegraphics[width=0.135\linewidth]{fig/qualitative_results/nyu/img_5.png} & 
        \includegraphics[width=0.135\linewidth]{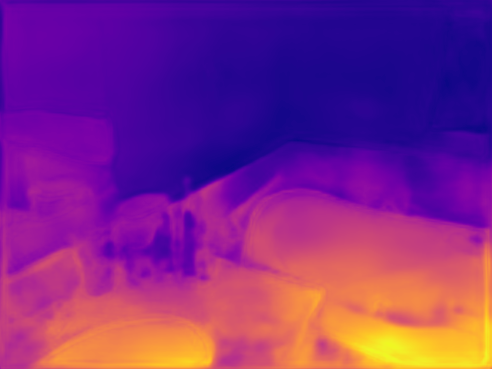} & 
        \includegraphics[width=0.135\linewidth]{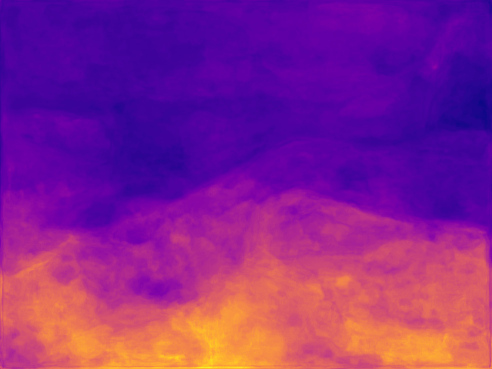} &
        \includegraphics[width=0.135\linewidth]{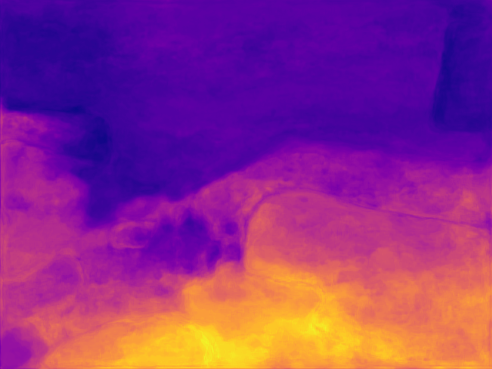} & 
        \includegraphics[width=0.135\linewidth]{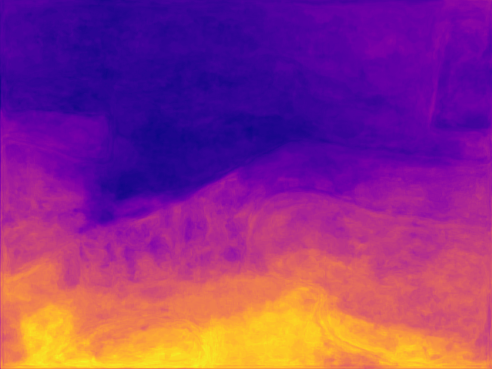} &
        \includegraphics[width=0.135\linewidth]{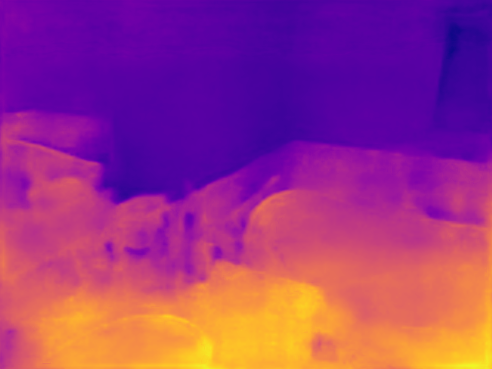} & 
        \includegraphics[width=0.135\linewidth]{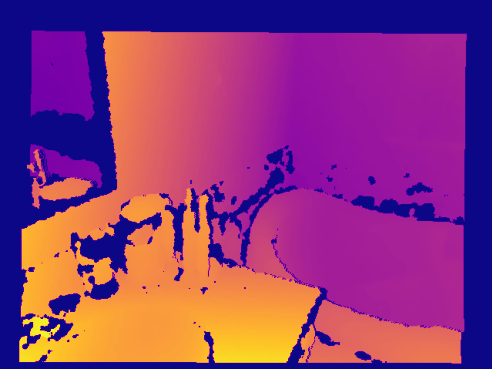}
        \\
        \includegraphics[width=0.135\linewidth]{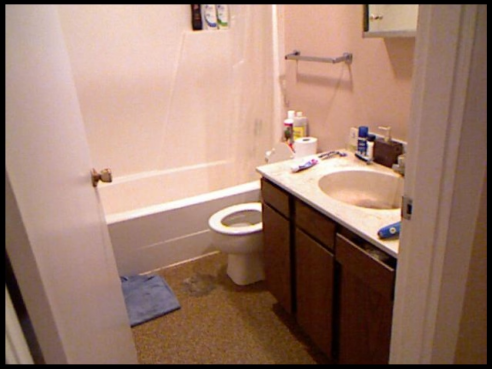} & 
        \includegraphics[width=0.135\linewidth]{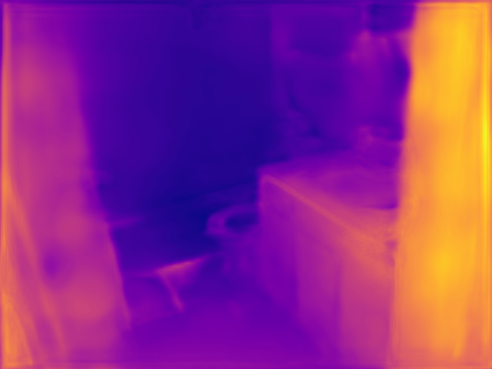} & 
        \includegraphics[width=0.135\linewidth]{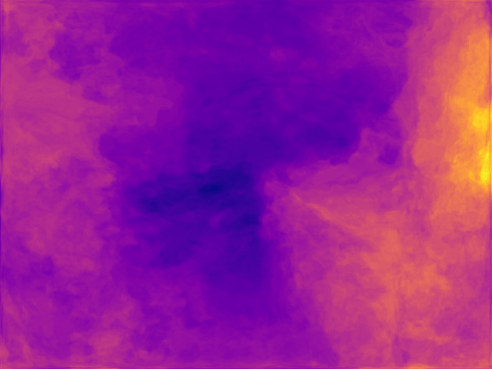} &
        \includegraphics[width=0.135\linewidth]{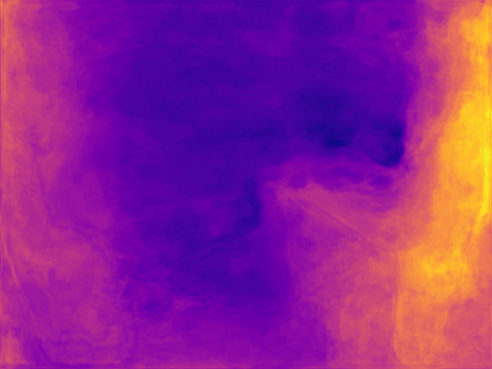} & 
        \includegraphics[width=0.135\linewidth]{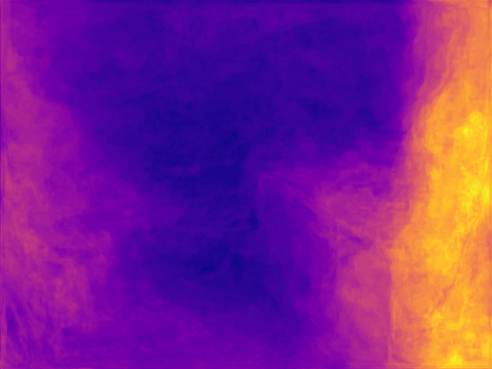} &
        \includegraphics[width=0.135\linewidth]{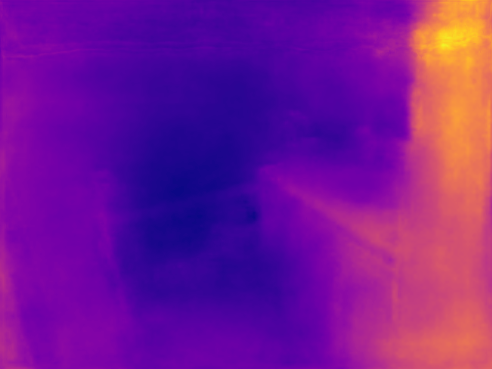} & 
        \includegraphics[width=0.135\linewidth]{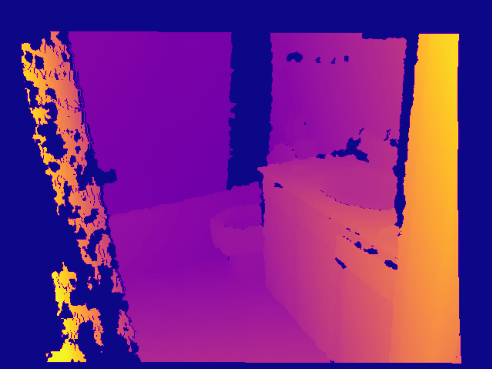}
        \\
        \includegraphics[width=0.135\linewidth]{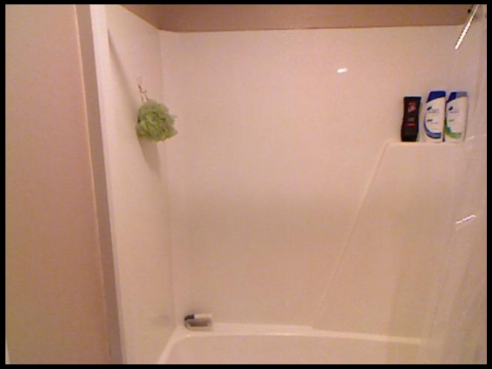} & 
        \includegraphics[width=0.135\linewidth]{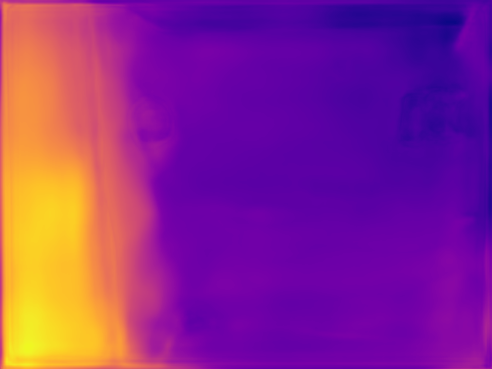} & 
        \includegraphics[width=0.135\linewidth]{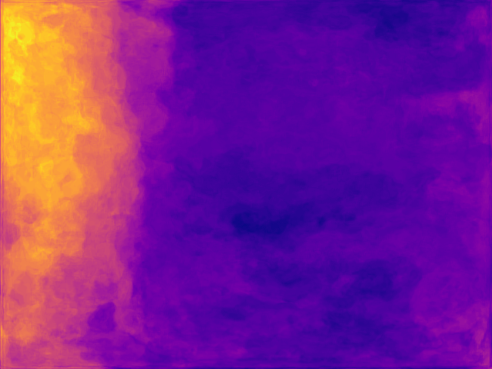} &
        \includegraphics[width=0.135\linewidth]{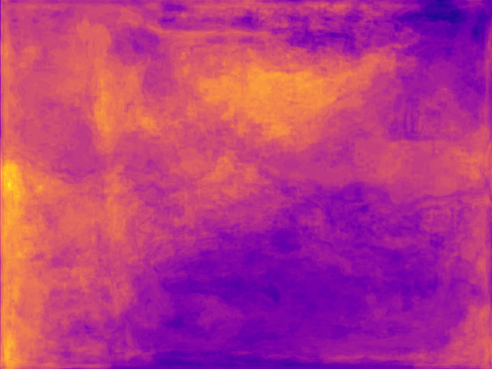} & 
        \includegraphics[width=0.135\linewidth]{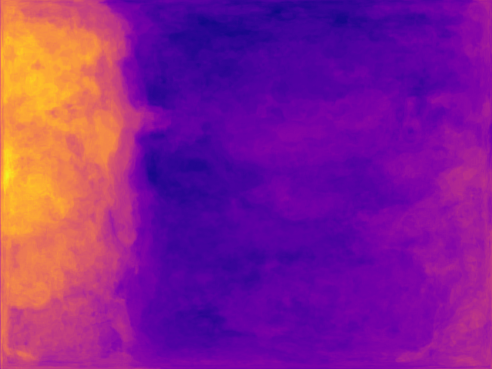} &
        \includegraphics[width=0.135\linewidth]{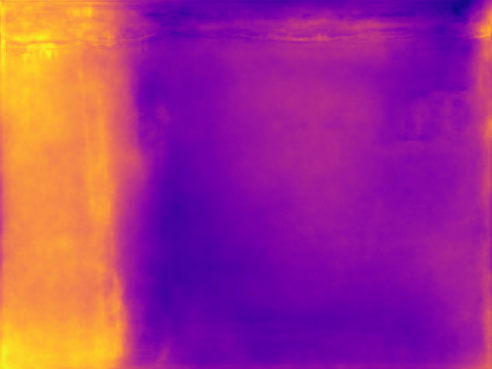} & 
        \includegraphics[width=0.135\linewidth]{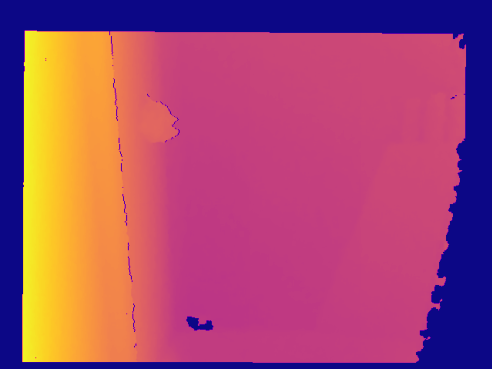}
    \end{tabular}

    \begin{tabular}{ccc}
         & ConvNeXt~\cite{liu2022convnet} & BiDense (Ours) \\
        \includegraphics[width=0.325\linewidth]{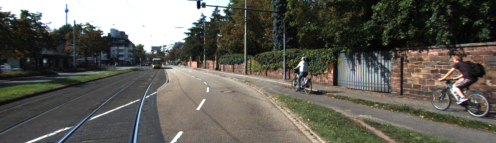} & 
        \includegraphics[width=0.325\linewidth]{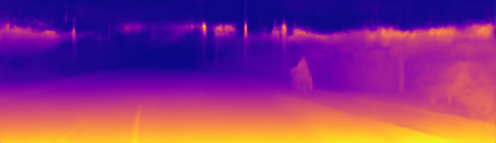} & 
        \includegraphics[width=0.325\linewidth]{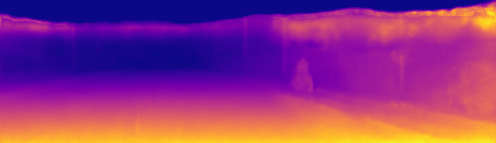}
        \\
        \includegraphics[width=0.325\linewidth]{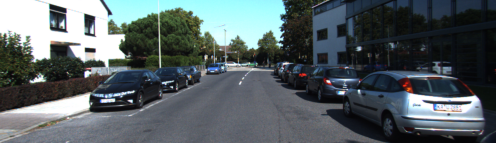} & 
        \includegraphics[width=0.325\linewidth]{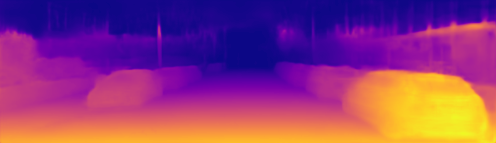} & 
        \includegraphics[width=0.325\linewidth]{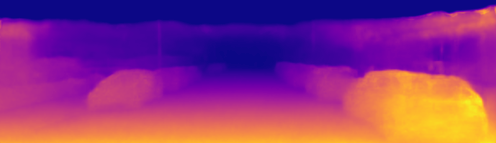}
        \\
        \includegraphics[width=0.325\linewidth]{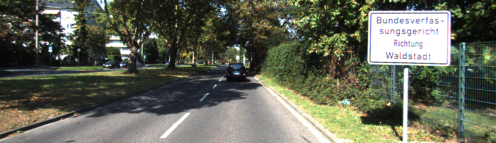} & 
        \includegraphics[width=0.325\linewidth]{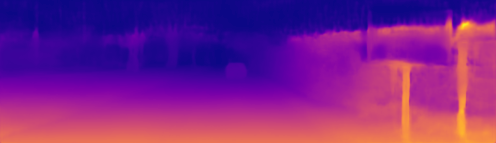} & 
        \includegraphics[width=0.325\linewidth]{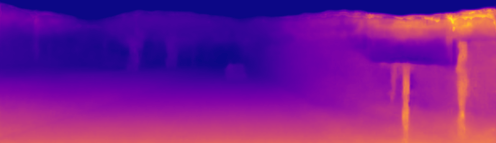}
        \\
        \includegraphics[width=0.325\linewidth]{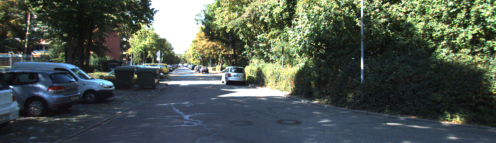} & 
        \includegraphics[width=0.325\linewidth]{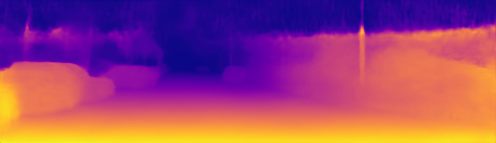} & 
        \includegraphics[width=0.325\linewidth]{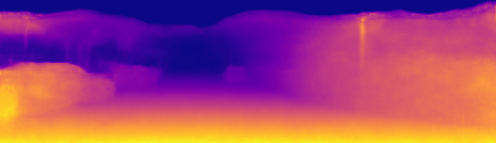}
        \\
        % \includegraphics[width=0.325\linewidth]{fig/qualitative_results/kitti/img_300.png} & 
        % \includegraphics[width=0.325\linewidth]{fig/qualitative_results/kitti/fp32_0_300.png} & 
        % \includegraphics[width=0.325\linewidth]{fig/qualitative_results/kitti/bidense_300.png}
        % \\
        \includegraphics[width=0.325\linewidth]{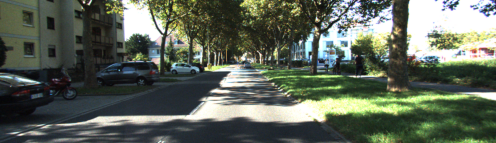} & 
        \includegraphics[width=0.325\linewidth]{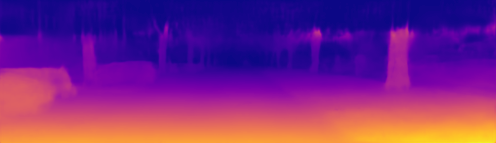} & 
        \includegraphics[width=0.325\linewidth]{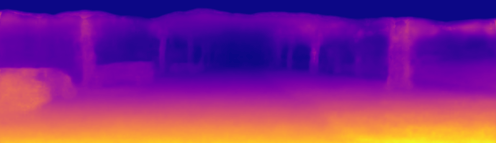}
        \\
        % \includegraphics[width=0.325\linewidth]{fig/qualitative_results/kitti/img_500.png} & 
        % \includegraphics[width=0.325\linewidth]{fig/qualitative_results/kitti/fp32_0_500.png} & 
        % \includegraphics[width=0.325\linewidth]{fig/qualitative_results/kitti/bidense_500.png}
        % \\
        \includegraphics[width=0.325\linewidth]{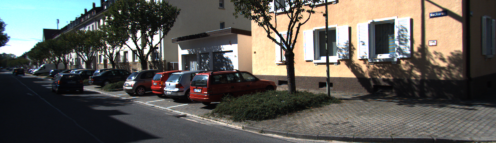} & 
        \includegraphics[width=0.325\linewidth]{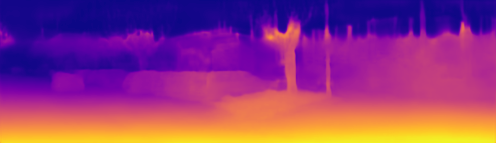} & 
        \includegraphics[width=0.325\linewidth]{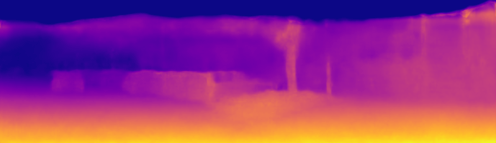}
    \end{tabular}
    \vspace{-2mm}
    \caption{Qualitative results of monocular depth estimation results on the NYUv2~\cite{silberman2012indoor} and KITTI~\cite{geiger2012we} depth datasets.}
    \label{fig:add-nyu}
    \vspace{-1mm}
\end{figure*}

\begin{figure*}[htbp]
    \centering
    \footnotesize
    \tabcolsep=1.5pt
    \begin{tabular}{ccccccc}
        ConvNeXt~\cite{liu2022convnet} & BNN~\cite{hubara2016binarized} & ReActNet~\cite{liu2020reactnet} & AdaBin~\cite{tu2022adabin} & BiSRNet~\cite{cai2024binarized} & BiDense (Ours) & Ground Truth\\
        \includegraphics[width=0.135\linewidth]{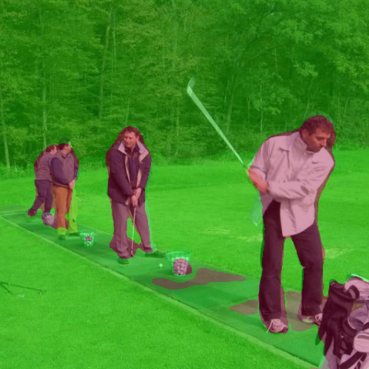} & 
        \includegraphics[width=0.135\linewidth]{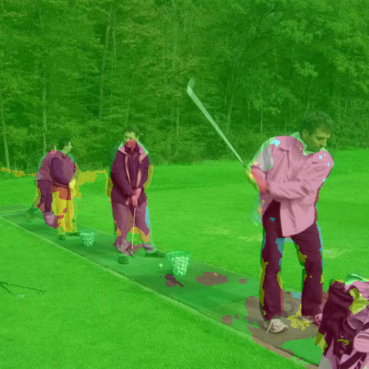} & 
        \includegraphics[width=0.135\linewidth]{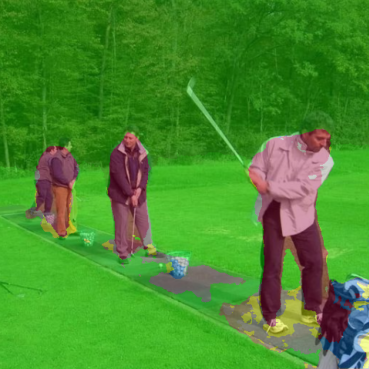} & 
        \includegraphics[width=0.135\linewidth]{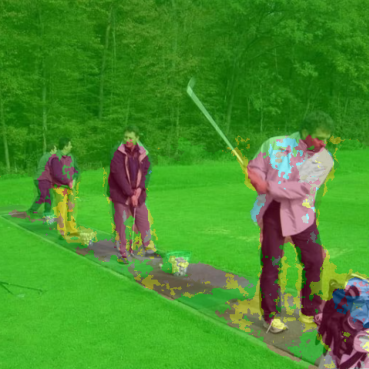} & 
        \includegraphics[width=0.135\linewidth]{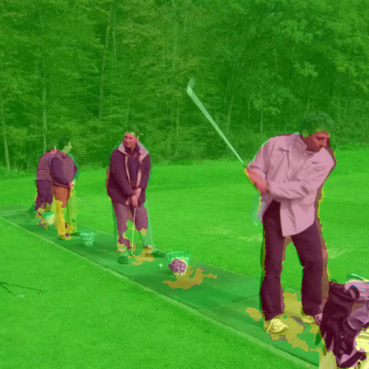} & 
        \includegraphics[width=0.135\linewidth]{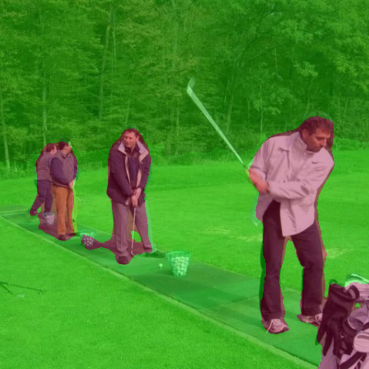} & 
        \includegraphics[width=0.135\linewidth]{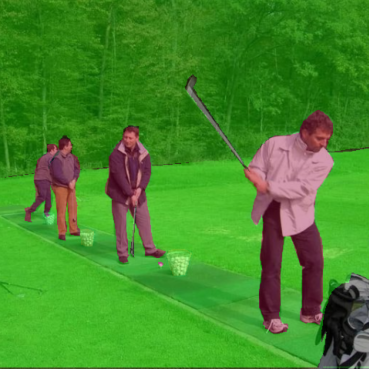} \\
        
        \includegraphics[width=0.135\linewidth]{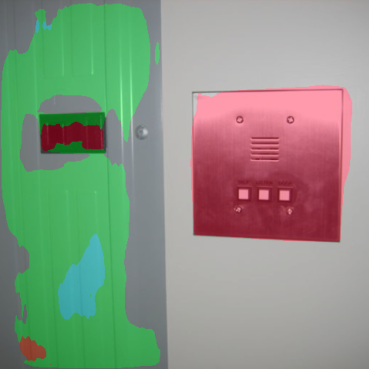} & 
        \includegraphics[width=0.135\linewidth]{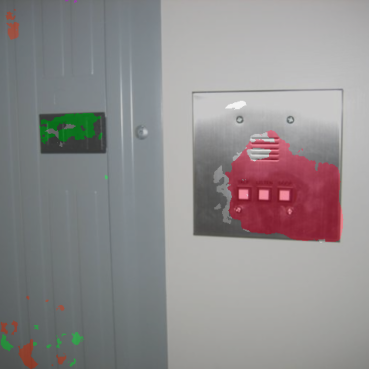} & 
        \includegraphics[width=0.135\linewidth]{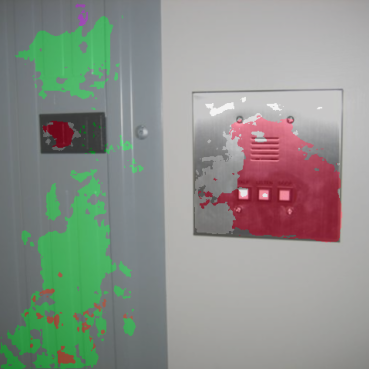} & 
        \includegraphics[width=0.135\linewidth]{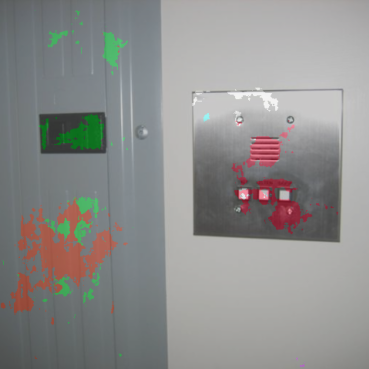} & 
        \includegraphics[width=0.135\linewidth]{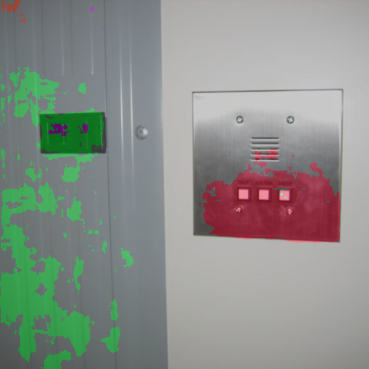} & 
        \includegraphics[width=0.135\linewidth]{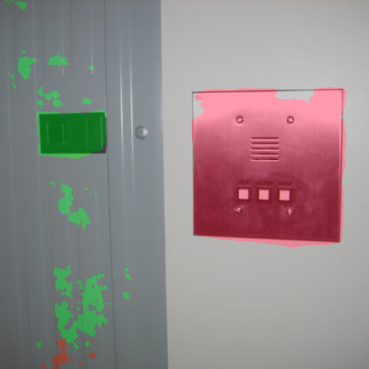} & 
        \includegraphics[width=0.135\linewidth]{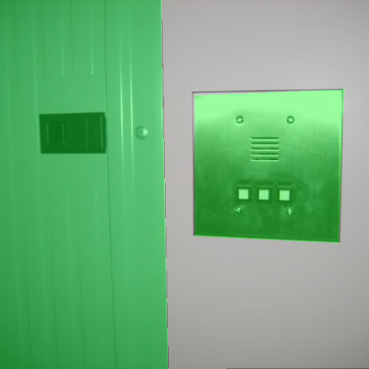} \\
        
        \includegraphics[width=0.135\linewidth]{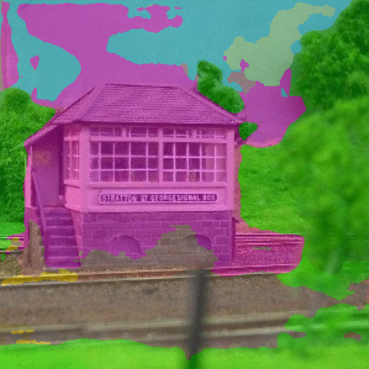} & 
        \includegraphics[width=0.135\linewidth]{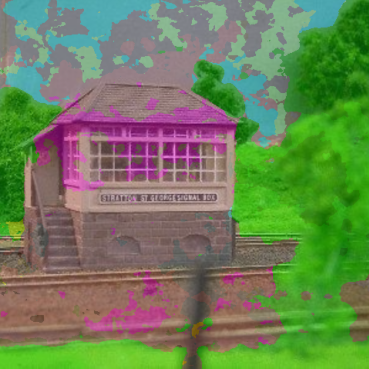} & 
        \includegraphics[width=0.135\linewidth]{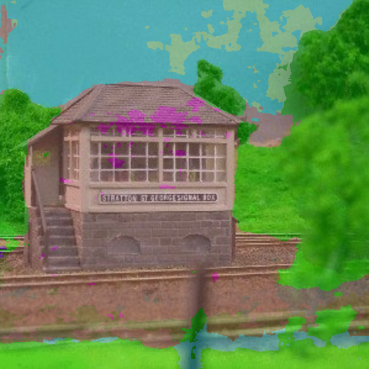} & 
        \includegraphics[width=0.135\linewidth]{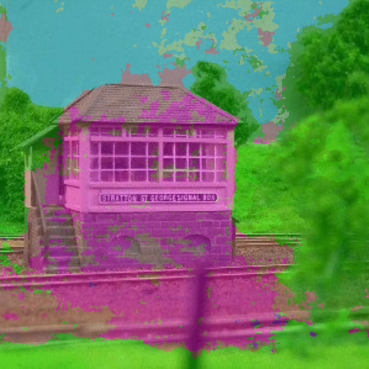} & 
        \includegraphics[width=0.135\linewidth]{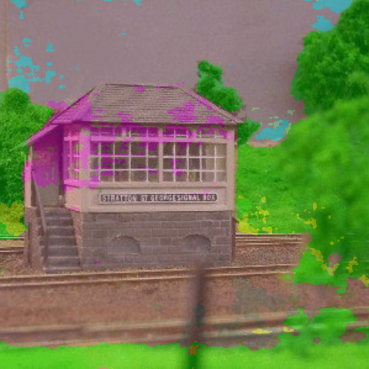} & 
        \includegraphics[width=0.135\linewidth]{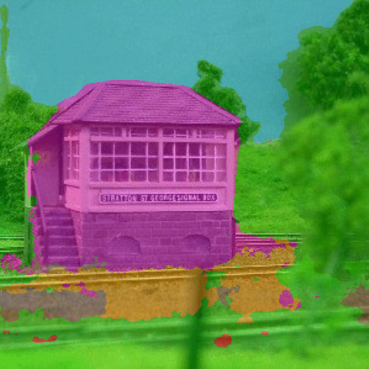} & 
        \includegraphics[width=0.135\linewidth]{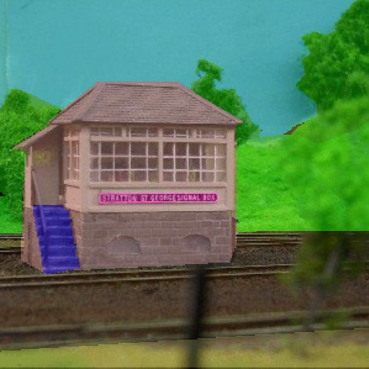} \\
        
        \includegraphics[width=0.135\linewidth]{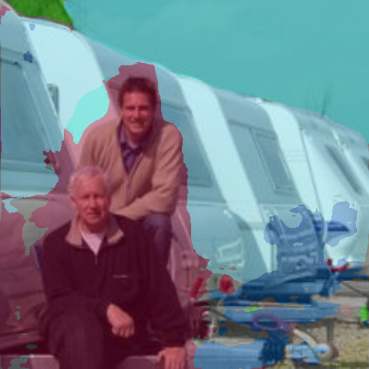} & 
        \includegraphics[width=0.135\linewidth]{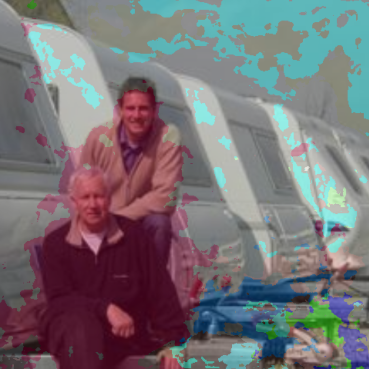} & 
        \includegraphics[width=0.135\linewidth]{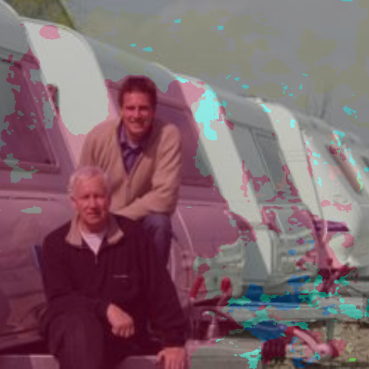} & 
        \includegraphics[width=0.135\linewidth]{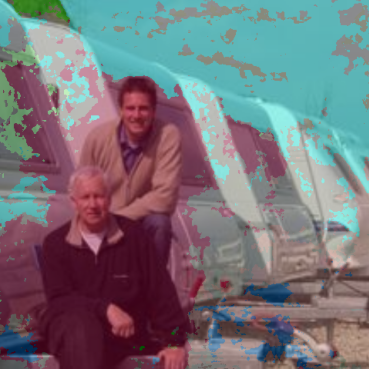} & 
        \includegraphics[width=0.135\linewidth]{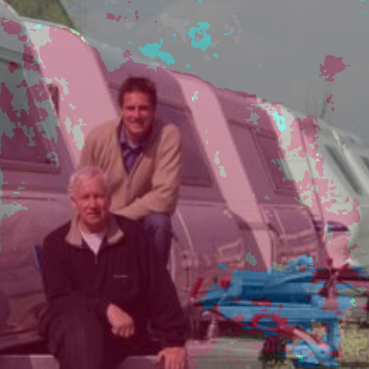} & 
        \includegraphics[width=0.135\linewidth]{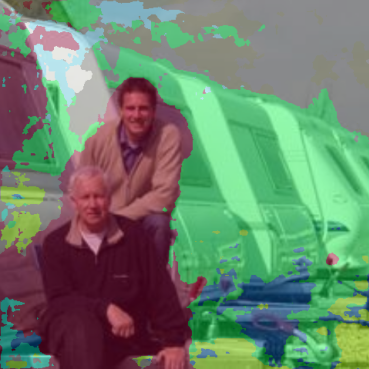} & 
        \includegraphics[width=0.135\linewidth]{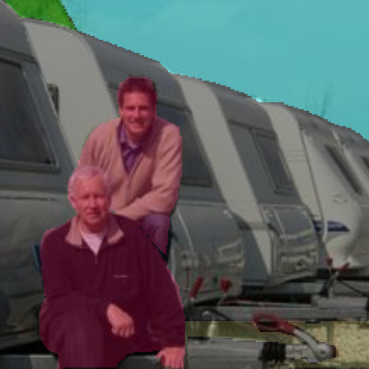} \\

        \includegraphics[width=0.135\linewidth]{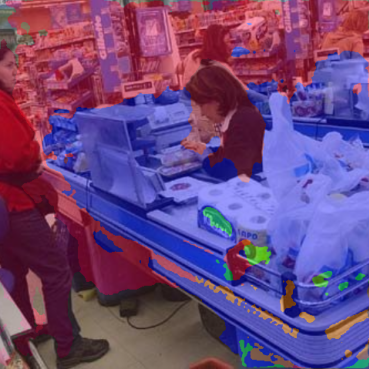} & 
        \includegraphics[width=0.135\linewidth]{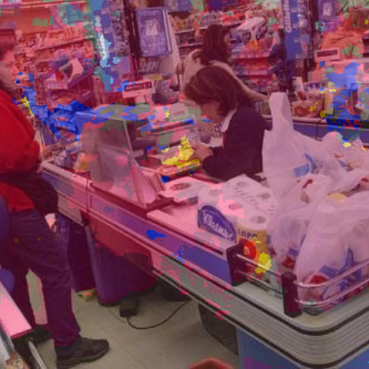} & 
        \includegraphics[width=0.135\linewidth]{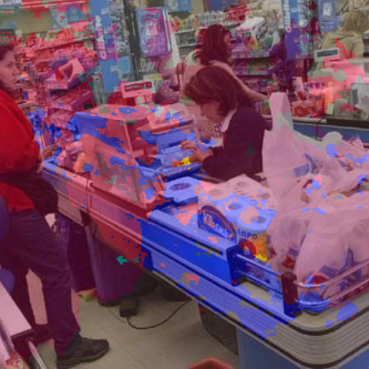} & 
        \includegraphics[width=0.135\linewidth]{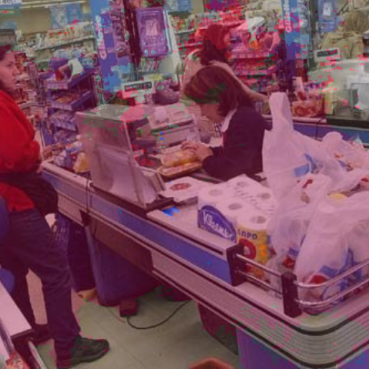} & 
        \includegraphics[width=0.135\linewidth]{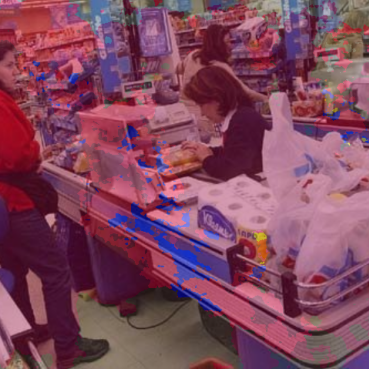} & 
        \includegraphics[width=0.135\linewidth]{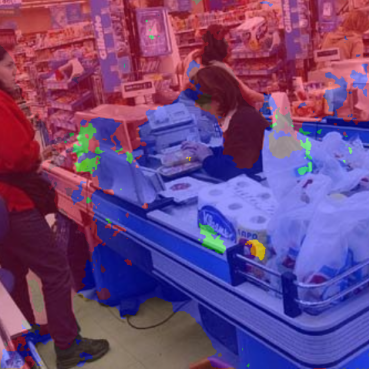} & 
        \includegraphics[width=0.135\linewidth]{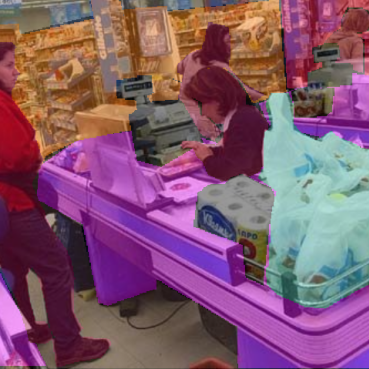} \\

        \includegraphics[width=0.135\linewidth]{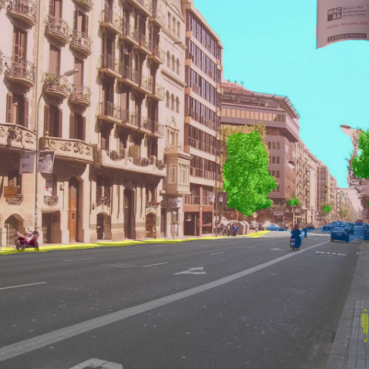} & 
        \includegraphics[width=0.135\linewidth]{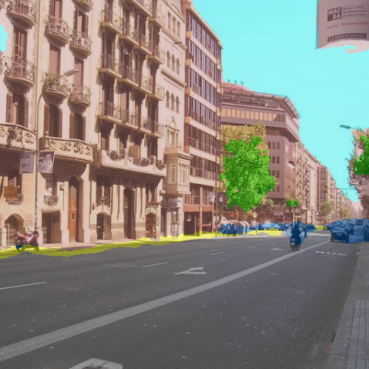} & 
        \includegraphics[width=0.135\linewidth]{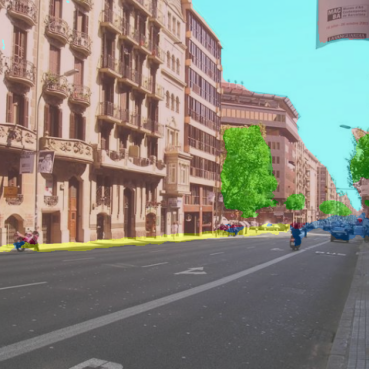} & 
        \includegraphics[width=0.135\linewidth]{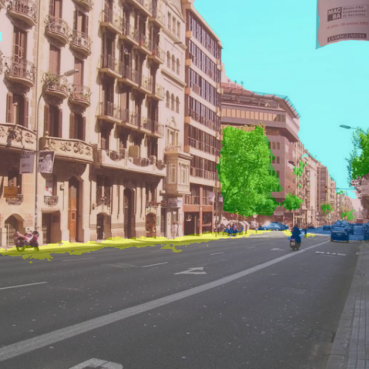} & 
        \includegraphics[width=0.135\linewidth]{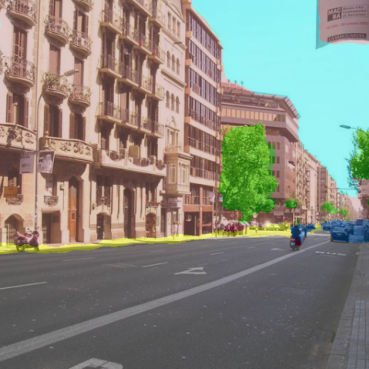} & 
        \includegraphics[width=0.135\linewidth]{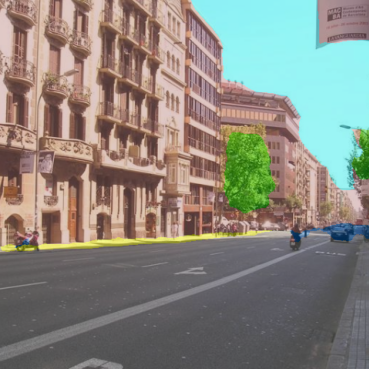} & 
        \includegraphics[width=0.135\linewidth]{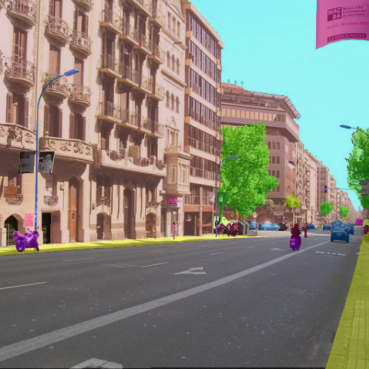} \\

    \end{tabular}
     \vspace{-1mm}
    \caption{Qualitative Results of semantic segmentation on the ADE20K~\cite{zhou2017scene} validation set.}
    \label{fig:add-ade20k-viz}
    % \vspace{-3mm}
\end{figure*}

\end{document}